\documentclass[journal]{IEEEtran}
\usepackage[utf8]{inputenc}

\ifCLASSINFOpdf
  
\else

\fi

\usepackage{url}
\usepackage{amsthm}
\usepackage{afterpage}
\usepackage{adjustbox}
\usepackage{setspace}
\usepackage[normalem]{ulem}
\useunder{\uline}{\ul}{}
\usepackage{graphicx}
\usepackage{textcomp}
\usepackage{xcolor}
\usepackage{balance} 
\usepackage{bbm}
\usepackage{amsmath,amsfonts}
\usepackage{epstopdf}
\usepackage{array}
\usepackage{tabularx}
\usepackage{booktabs}
\usepackage{color}
\usepackage{xcolor}
\usepackage{colortbl}
\usepackage{xspace}
\usepackage{multirow}
\usepackage{pifont}
\usepackage{enumitem}
\usepackage{bbding}
\usepackage{hyperref}
\usepackage{makecell}
\usepackage{microtype}
\usepackage{fontawesome}
\usepackage{caption,setspace}
\usepackage{subfigure}
\usepackage[utf8]{inputenc}
\usepackage{utfsym}
\usepackage{diagbox}
\usepackage{makecell}
\usepackage[linesnumbered,ruled,vlined]{algorithm2e}

\newcommand{\ssymbol}[1]{^{\@fnsymbol{#1}}}

\SetKwInput{KwInput}{Input}                
\SetKwInput{KwOutput}{Output}              
\SetKwRepeat{Do}{do}{while}

\definecolor{darkred}{rgb}{0.55, 0.0, 0.0}
\definecolor{lightred}{rgb}{0.94, 0.5, 0.5}
\definecolor{rosybrown}{rgb}{0.74, 0.56, 0.56}
\definecolor{darkgreen}{rgb}{0.0, 0.39, 0.0}
\definecolor{skyblue}{rgb}{0.56, 0.93, 0.56}
\definecolor{darkblue}{rgb}{0.0, 0.0, 0.55}
\definecolor{lightblue}{rgb}{0.68, 0.85, 0.9}
\definecolor{skyblue}{rgb}{0.53, 0.81, 0.92}

\definecolor{gray}{rgb}{0.5, 0.5, 0.5}
\definecolor{forestgreen}{rgb}{0.13, 0.55, 0.13}
\definecolor{slateblue}{rgb}{0.42, 0.35, 0.80}
\definecolor{darkgray}{rgb}{0.33, 0.33, 0.33}
\definecolor{royalblue}{rgb}{0.25, 0.41, 0.88}

\SetCommentSty{mycommfont}

\usepackage{ulem}

\begin{document}

\title{A Comprehensive Data-centric Overview of Federated Graph Learning}

\author{Zhengyu Wu, Xunkai Li, Yinlin Zhu, Zekai Chen, Guochen Yan, Yanyu Yan, \\ Hao Zhang, Yuming Ai, Xinmo Jin, Rong-Hua Li, Guoren Wang
\thanks{Zhengyu Wu, Xunkai Li, Zekai Chen, Yuming Ai, Xinmo Jin, Rong-Hua Li, and Guoren Wang are with Beijing Institute of Technology, Beijing, 100811, China. (e-mail: jeremywzy96@outlook.com; cs.xunkai.li@gmail.com; Zekai96@outlook.com; 1120211787@bit.edu.cn; 1287703522@qq.com; lironghuabit@126.com; wanggrbit@gmail.com) Yinlin Zhu is with Sun Yat-sen University, Guangzhou, 510006, China. (e-mail:zhuylin27@mail2.sysu.edu.cn) Guochen Yan is with Peking University, Beijing, 100871, China (e-mail:Guochen\_yan@outlook.com) Yanyu Yan is with Institute of Information Science, Beijing Jiaotong University, Beijing 100044, China. (e-mail:23111086@bjtu.edu.cn) Hao Zhang is with Computer Network Information Center, Chinese Academy of Sciences, Beijing, 100083, China. (email:hzhang@cnic.cn)}
}


\maketitle
\begin{abstract}
In the era of big data applications, Federated Graph Learning (FGL) has emerged as a prominent solution that reconcile the trade-off between optimizing the collective intelligence between decentralized datasets holders and preserving sensitive information to maximum. 
Existing FGL surveys have contributed meaningfully but largely focus on integrating Federated Learning (FL) and Graph Machine Learning (GML), resulting in early-stage taxonomies that emphasis on methodology and simulated scenarios.
Notably, a \textbf{data-centric perspective}—which systematically examines FGL methods through the lens of data properties and usage—remains unadapted to reorganize FGL research, yet it is critical to assess how FGL studies manages to tackle data-centric constraints to enhance model performances. 
This survey propose a two-level data-centric taxonomy: \textbf{Data Characteristics}, which categorizes studies based on the structural and distributional properties of datasets used in FGL, and \textbf{Data Utilization}, which analyzes the training procedures and techniques employed to overcome key data-centric challenges.
Each taxonomy level is defined by three orthogonal criteria, each representing a distinct data-centric configuration.
Beyond taxonomy, this survey examines FGL’s integration with Pre-trained Large Models, showcases real-world applications, and highlights future directions aligned with emerging trends in GML.
\end{abstract}

\begin{IEEEkeywords}
Federated Graph Learning, Machine Learning
\end{IEEEkeywords}

\IEEEpeerreviewmaketitle

\section{Introduction}
\IEEEPARstart{G}{raph} datasets, structured as non-Euclidean representations, are formally defined as tuples of nodes (entities) and edges (relationships) to rigorously model complex real-world systems.
A key advantage of graph datasets is their ability to explicitly encode topological connections, overcoming the traditional constraints of independent and identically distributed (i.i.d.) data by capturing interaction dependencies among entities directly~\cite{non-iid}.
Unlike conventional data formats such as pixel-based images or text, graph structures offer distinct theoretical benefits, and the introduction of  Graph Neural Networks (GNNs) enables Machine Learning (ML) algorithms to explore implicit structural insights hidden within topological connections based on the propagation mechanism. 
By virtue of their effectiveness, GNNs have enabled breakthroughs such as AlphaFold~\cite{alaphafold}, which predicts protein structures from amino acid sequences to advance vaccine and antibody development.
 
Given the demonstrated efficacy of GNNs, numerous groundbreaking models, such as GCN~\cite{GCN} and GAT~\cite{GAT}, have been proposed. The majority of these approaches adopt a model-centric perspective that emphasizes on introducing innovative architectural designs to achieve optimal performance on given datasets.~\cite{data-centricsurvey}
However, their success depends on the implicit assumption that the datasets have been thoroughly refined, and that performance improvements primarily stem from increasingly sophisticated model architectures.
On the other hand, real-world data often exhibit significant uncertainties—such as undesirable noise and incomplete descriptions of extracted entities—that violate this assumption. When such low-quality datasets are fed into GNNs, the models struggle to efficiently extract reliable knowledge, thereby exposing the inherent vulnerability of model-centric approaches in practical applications.

To address these limitations, data-centric Graph Machine Learning (GML) has emerged as a leading paradigm that more pragmatically addresses real-world data challenges. Consequently, data-centric GML has garnered increasing attention from researchers. However, most studies assume centralized settings, where datasets are uniformly stored in a single location. In contrast, decentralized data-centric GML remains underexplored, despite the critical reality that data are often dispersed across multiple, independent holders.
Meanwhile, processing decentralized data requires stringent regulation compliance to preserve privacy. 
In response, Federated Learning (FL) has gained attention for it enabling collaborative model training across decentralized datasets while preserving privacy.~\cite{FederatedScope_LLM} 

Extending FL to graph datasets, Federated Graph Learning (FGL) has evolved swiftly as a specialized decentralized graph learning framework. 
Existing FGL studies share the pattern of proposing research challenges rooted in suggested realistic scenarios, based on which existing surveys have established taxonomies emphasizing prevalently adopted scenario-based challenges. 
Such a pursuit has contributed meaningfully to the development of the field, but the scope of which are originated from model-centric perspective, introducing the innovative mechanism without delineating differentiated properties of datasets nor discussing their data-centric motivations. 

\textit{\textbf{Motivation of this survey}:}
Inspiration for this data-centric FGL survey stems from the unequivocal understanding that most FGL challenges are data-related, such as statistical heterogeneity or topology heterogeneity. 
Moreover, examining those challenges closely requires the acknowledgment of data characteristics applied in FGL works since diversified data formats and decentralized settings have been discussed in existing works. 
The correspondent mechanism is also knitted with data-centric GML but in decentralized settings. 
To better present FGL as a general research field to researchers whose initial intention is to resolve observed data-related issues, having the data-centric FGL survey can offer greater convenience as a generalized guide.

Specifically, we introduce a two-level taxonomy, each comprising three orthogonal criteria that form unique combinations offering comprehensive view: 

    The \textbf{Data Characteristics Dimension} includes: (i) differentiating various types of graph datasets (e.g., homogeneous, heterogeneous, knowledge, and bipartite graphs), (ii) highlighting the decentralization format describing how data is distributed across clients, and (iii) revealing the visibility level at each client, indicating whether each client has access to a full global graph or only partial subgraphs. Together, these criteria provide a comprehensive view of the structural and distributional properties of data addressed in FGL research. 
    
    The \textbf{Data Utilization Dimension}, in turn, examines how and when FGL methods incorporate targeted mechanisms that aim for addressing data-centric challenges into the training process: (i) highlighting key data-centric challenges, including impaired data quality, class imbalance across clients' datasets, slow convergence when training on excessively large graphs, and enhancement of data privacy protection. (ii) indication of whether main innovations are proposed in client-side or server-side procedure. (iii) further refining the training procedure into four execution phases (e.g, Initialization, Local Training, Global Aggregation, and Post-aggregation) with details of techniques applied in representative FGL methods.

As the first FGL survey focusing on data-centric perspective, the contribution of this work can be illustrated as follows: \\
\noindent
\textbf{\textit{(a) New Perspective}}: This work is the first to organize FGL studies through a data-centric lens, clarifying how different types of data are characterized and utilized in existing research. This perspective aligns closely with the priorities of the big data era, where the properties of data increasingly shape the choice and effectiveness of machine learning techniques. \\
\noindent
\textbf{\textit{ (b) Two-level Taxonomy}}: We propose a new two-level taxonomy grounded in a data-centric perspective. 
Each level of taxonomy comprises three orthogonal criteria that categorize all notable FGL studies in a fine-grained fashion and facilitating the discovery of studies aligned with specific data-centric questions.\\
\noindent
\textbf{\textit{ (c) Broader Impact-Artificial Generative Intelligence}}: This work is also the first to explore how FGL could integrate with ongoing research on Pre-trained Large Models (PLMs) to accelerate advances in graph machine learning. Our discussion of future research directions highlights several emerging data-centric topics that remain underexplored in the context of FGL and thus merit further attention and dedicated investigation.

\noindent
\textit{\textbf{Organization of the Survey}}
The structure of this survey is as follows: Sec.\ref{sec: preliminaries} introduces the foundational concepts of FL and FGL, outlining the general training procedure. Sec.\ref{sec: data characteristics} presents the first-level taxonomy based on data properties from both local and global perspectives. Sec.\ref{sec: data utilization} lays out second-level Taxonomy, which presents data-centric challenges and details how representative FGL methods respond to them. Sec.\ref{sec: euclidean-oriented FGL} discusses studies where clients operate on non-graph-structured data. Sec.\ref{sec: Data Application} evaluates FGL’s applicability in addressing real-world data challenges. Sec.\ref{sec: PLM} explores the mutual integration of FGL and PLMs. Finally, Sec.~\ref{sec: future works} outlines future directions, exploring FGL's integration with trending GML topics and extending FGL to more topologically complex graph types.
 
\section{Background and Terminology}
\label{sec: preliminaries}

\subsection{Data-centric Perspective}
Data is the foundation of artificial intelligence (AI) and machine learning, with its role in model performance becoming increasingly central. Yet, progress is often limited by two core challenges: poor data quality (\textit{quality issues}) and the need for large, diverse datasets (\textit{quantity issues}). These concerns have made data a focal point in the rise of data-centric AI.
Unlike the traditional model-centric paradigm—which prioritizes architectural design while keeping data fixed—the data-centric approach addresses issues inherent to the training data itself. It shifts the emphasis from model innovation to improving data quality and utility.
Specifically, data-centric AI seeks to overcome limitations such as missing or incorrect labels, structural noise (e.g., faulty or absent edges), and incomplete graph data (e.g., missing neighbors). By enhancing data integrity and completeness, this paradigm aims to support more robust and generalizable AI systems.

FGL was introduced to enable privacy-preserving distributed learning across multiple parties. While this goal remains central, FGL also presents unique data-centric challenges.
To preserve local data, FGL simulations typically rely on graph partitioning strategies (namely Metis~\cite{karypis1998metis} and Louvain~\cite{blondel2008louvain}) adapted from community detection to divide a global graph into local subgraphs. However, this partitioning fragments the graph’s structure, leading to data heterogeneity across clients.
This heterogeneity is commonly categorized as statistical and topological. While statistical heterogeneity, such as skews in node features or label distributions, is well-studied in traditional FL, its effects are more intricate in graph settings due to entanglement with the underlying topology.
More recent works~\cite{li2024adafgl, li2024fedgta, zhu2024fedtad} have shifted attention to topological heterogeneity, emphasizing its role in hindering learning. In particular, dominant heterophilous connections within local subgraphs can degrade model performance, highlighting the necessity to mitigate structural inconsistencies across clients.

To holistically present data-centric challenges in FGL, we reframe them from a broader perspective.
Rather than adhering strictly to conventional categories like statistical and topological heterogeneity, this survey proposes a taxonomy comprising two orthogonal dimensions: \textbf{Data Characteristics} and \textbf{Data Utilization}.
Data Characteristics (Sec.~\ref{sec: data characteristics}) captures variations introduced during the simulation phase, including the data type, partitioning strategy, and properties of the resulting local subgraphs.
Data Utilization (Sec.~\ref{sec: data utilization}) focuses on methodologies designed to address data-centric issues such as low data quality, limited cross-client collaboration, and privacy risks. 
Such explicit categorizing is essential for researchers to acquire the comprehensive understanding on FGL with clear connections between data-centric motivation and method designs.

As this survey adopts a data-centric perspective, we present a brief description of the training procedure, which best categorize the essence of FGL training, in a model-agonistic fashion.
Specifically, we present a representative setup using  GCN as the backbone and FedAvg as the default aggregation strategy. This serves as a conceptual foundation for understanding the FGL pipeline from a data-centric standpoint.

\subsection{Graph Neural Networks (GNNs)}
GNNs are a class of deep learning models designed to operate on graph-structured data, utilizing both the graph's topology and node features to learn representations. A graph \( G = (A, X, Y) \) consists of the adjacency matrix \( A \), which encodes the connectivity between nodes, the node feature matrix \( X \), which represents node attributes, and the node labels \( Y \), used for supervised learning tasks. GNNs employ a message-passing process where node features are iteratively aggregated based on neighboring nodes.

Taking the GCN~\cite{kipf2016gcn} as an example, in which the propagation of information in the \( \ell \)-th layer is given by:
\begin{equation}
H^{(\ell)} = \text{ReLU}\left(\hat{A} H^{(\ell-1)} W^{(\ell)}\right),
\end{equation}
where \( H^{(0)} = X \) is the initial node feature matrix, \( \hat{A} \) is the normalized adjacency matrix with self-loops, and \( W^{(\ell)} \) is the trainable weight matrix.

After \( L \) layers, GNNs capture \( L \)-hop neighborhood information, and the final node embeddings are represented as:
\begin{equation}
H = f(A, X),
\end{equation}
where \( f \) is the GNN model. These embeddings \( h_i^{(L)} \) are then passed to a classifier \( F \) for task-specific predictions:
\begin{equation}
z_i = F(h_i).
\end{equation}
\subsection{Federated Learning (FL)}
In FL, each client \( c_k \) holds a private dataset \( D_k = \{(x_i, y_i)\}_{i=1}^{N_k} \), with \( x_i \) as features and \( y_i \) as the label for the \( i \)-th sample, and \( N_k \) being the number of samples on client \( c_k \). The objective of FL is to optimize the global model while maintaining privacy, defined as:
\begin{equation}
\min_{\omega} \frac{1}{N} \sum_{k=1}^{M} N_k L_k(\omega) = \min_{\omega} \frac{1}{N} \sum_{k=1}^{M} \sum_{i=1}^{N_k} l_k(x_i, y_i; \omega),
\end{equation}
where \( L_k(\omega) \) is the average loss on client \( c_k \), and \( N \) is the total number of samples across all clients.

The model training involves a central server aggregating local model updates from clients, typically using FedAvg \cite{mcmahan2017fedavg}:

\begin{equation}
    w_r = \frac{1}{n} \sum_{c=1}^{C} n_c w_r^c,
\end{equation}
where \( w_r^c \) and \( w_r \) represent the local and global model parameters at round \( r \), respectively, and \( n_c \) is the number of samples at client \( c \). 
\subsection{Federated Graph Learning (FGL)}
FGL extends FL to graph-structured data, enabling each client to independently train a GNN on their local graphs while preserving privacy. For a node classification task, the optimization objective in FGL for a client \( i \) can be written as:
\begin{equation}
\min_{\theta} \frac{1}{N} \sum_{i=1}^N \frac{|V^i|}{|V|} \mathcal{L}(f_{\theta}(V^i), Y^i),
\end{equation}
where \( V^i \) represents the set of nodes owned by client \( i \), \( Y^i \) are the corresponding labels, \( f_{\theta} \) is the GNN model parameterized by \( \theta \), and \( \mathcal{L} \) denotes the task-specific loss function.

Once the local training is complete, the model parameters are sent to a central server. 
The server performs parameter aggregation by considering the number of nodes in each client’s graph. Specifically, the global model parameters \( w_t \) at time \( t \) are updated using a weighted average as follows:
\begin{equation}
w_t = \frac{\sum_{k \in S_t} |V_k|}{N} \sum_{k \in S_t} w_k^{(t)},
\end{equation}
where \( S_t \) denotes the set of clients participating in the aggregation at time \( t \), \( w_k^{(t)} \) is the model parameter set from client \( k \) at iteration \( t \), \( |V_k| \) is the number of nodes in the graph owned by client \( k \), and \( N \) represents the total number of nodes across all selected clients. This aggregated global model is then redistributed to the clients for further training rounds. Other GNN models, such as GAT~\cite{GAT}, GraphSage~\cite{hamilton2017graphsage}, also have been utilized in FGL studies, and for more comprehensive details of GNN models and their applications in FGL can be found in following surveys.~\cite{Zhou2020GNNsurvey, wu2020comprehensive}

\section{Comparison with other FGL survey}

Although three prior surveys have organized FGL studies into technical taxonomies that help researchers understand foundational concepts, their scope is limited by available studies at the time, and their taxonomies emphasize deployment scenarios rather than the core data challenges of FGL methods.

At the inception of FGL research, \cite{zhang2021fgl_survey_2} provided the initial FGL taxonomies that categorize approaches based on distribution of graph data into inter-graph FL, intra-graph FL (under the horizontal and vertical partition), and graph-structured FL. It focuses on defining each type, outlining application scenarios, and identifying key challenges. Based on this foundation, ~\cite{fu2022fgl_survey_1} adopt the same categorization scheme, applying it to the analysis of FGL training architectures. 
Its taxonomy emphasizes scenario-driven challenges prevalent in early FGL studies and differentiates the technical methods proposed to address them.
\cite{liu2025fgl_survey_3} clarify how FL and graph learning interact and complement each other. The taxonomy remains technically oriented, similar to that of \cite{fu2022fgl_survey_1}, but expands its scope to include additional scenarios and a broader range of studies.

By reorganizing the extensive body of FGL research from a data-centric perspective, this survey distinguishes itself from previous surveys in several key aspects as detailed below. Differences between this survey and others is listed in Table~\ref{tab: differences}. \\
(1) \textbf{Extending Data-Centric Distinctions} This work highlights variations in data formats across existing FGL studies and emphasizes how these differences give rise to distinct data-centric challenges. For instance, knowledge graphs differ significantly from conventional heterogeneous graphs in terms of node diversity and the semantic richness of relational structures.
The embedding techniques used to extract latent information from knowledge graphs often require distinctive learning procedures.
By explicitly categorizing these differences,
\begin{table}[!htbp]
    \centering
    \caption{A comparison between existing FGL surveys.}
    \scalebox{0.7}{
\setlength{\tabcolsep}{1mm}{
\begin{tabular}{c|ccc|c}
\toprule
FGL Surveys              & \begin{tabular}[c]{@{}c@{}}Graph\\ Properties\end{tabular} & \begin{tabular}[c]{@{}c@{}}Data-oriented\\ Challenges\end{tabular} & \begin{tabular}[c]{@{}c@{}} Technical\\ Discussion\end{tabular} & \begin{tabular}[c]{@{}c@{}}Included Works\\ Volume\end{tabular}\\ \midrule[0.3pt]
\textit{Zhang et al.} ~\cite{zhang2021fgl_survey_2}                & \textcolor{red}{\usym{2717}}             & \textcolor{red}{\usym{2717}}           & \textcolor{red}{\usym{2717}}              & 7 FGL works \\
\textit{Liu et al.} ~\cite{liu2025fgl_survey_3}            & \textcolor{red}{\usym{2717}}             & \textcolor{red}{\usym{2717}}           & \textcolor{teal}{\usym{2713}}             & 37 FGL works \\
\textit{Fu et al.} ~\cite{fu2022fgl_survey_1}               & \textcolor{red}{\usym{2717}}             & \textcolor{teal}{\usym{2713}}           & \textcolor{teal}{\usym{2713}}             & 61 FGL works \\
Data-centric FGL Survey        & \textcolor{teal}{\usym{2713}}             & \textcolor{teal}{\usym{2713}}    & \textcolor{teal}{\usym{2713}} & 79 FGL works \\ \bottomrule
\end{tabular}
    \label{tab: differences}}}
    \end{table}
this survey highlights the implications that data format has on learning behavior and system design in FGL. \\ 
(2) \textbf{Emphasizing Data-Centric Motivations} We observe that most FGL studies are closely tied to specific real-world scenarios, and the motivations for these studies often stem from particular data-centric concerns. For example, scenarios characterized by missing linkages naturally give rise to pronounced statistical heterogeneity. To offer a structured and accessible overview of these motivations, we classify existing work to align with broadly recognized data-centric challenges in AI. This taxonomy is especially useful for researchers less familiar with the FGL domain, enabling them to identify relevant studies aligned with their own data-centric goals. \\
(3) \textbf{Reframing Technical Contributions} Rather than organizing FGL techniques around application-specific scenarios, this survey examines when (i.e., in which phase of training) and where (i.e., on the client side, server side, or both) each study introduces its methodological innovations. This reframing provides a clearer and more systematic understanding of how different techniques address data-centric challenges and how they relate to the architectural components of FGL systems. 
\\
(4) \textbf{Integration between FGL and PLMs}  we explore the potential integration of FGL with Pre-trained Large Models (PLMs), reviewing current efforts that combine federated learning or centralized graph learning with PLMs. Building on these developments, we offer insights into how FGL, which is the convergence of both paradigms, can contribute to the next generation of PLMs research.
Through this data-centric lens, the survey not only complements existing technical overviews but also provides a practical roadmap for researchers and practitioners seeking to design robust and adaptive FGL systems that explicitly address data-centric concerns.

\section{Data Characteristics}
\label{sec: data characteristics}
In the first level of taxonomy, we present a systematic classification of FGL studies based on Data Characteristics, defined along three orthogonal dimensions: (1) graph datasets format (Sec.~\ref{subsec: data format}), (2) decentralization strategies in FGL simulations (Sec.~\ref{subsec: decentralization format}), and (3) data visibility at each client (Sec.~\ref{subsec: data combinations}). We then categorize existing FGL studies into combinations of three criteria and analyze the associated data properties. Given that aspects like decentralization format and client visibility have been partially discussed in prior surveys, this level functions as a complementary classification, offering a more integrative view. Accordingly, rather than detailing each method, we summarize them in Table~\ref{tab: data characteristics}, which provides a comprehensive mapping of existing works across the taxonomy.
Fig.~\ref{tab: data characteristics} displays visual illustrations of three criteria discussed below with realistic examples, such as patients records, citation networks, and e-commerce purchasing and rating system.   
\begin{figure*}[t]   
	\centering
    \setlength{\abovecaptionskip}{0.3cm}
    \setlength{\belowcaptionskip}{-0.3cm}
    \includegraphics[width=\linewidth,scale=1.00]{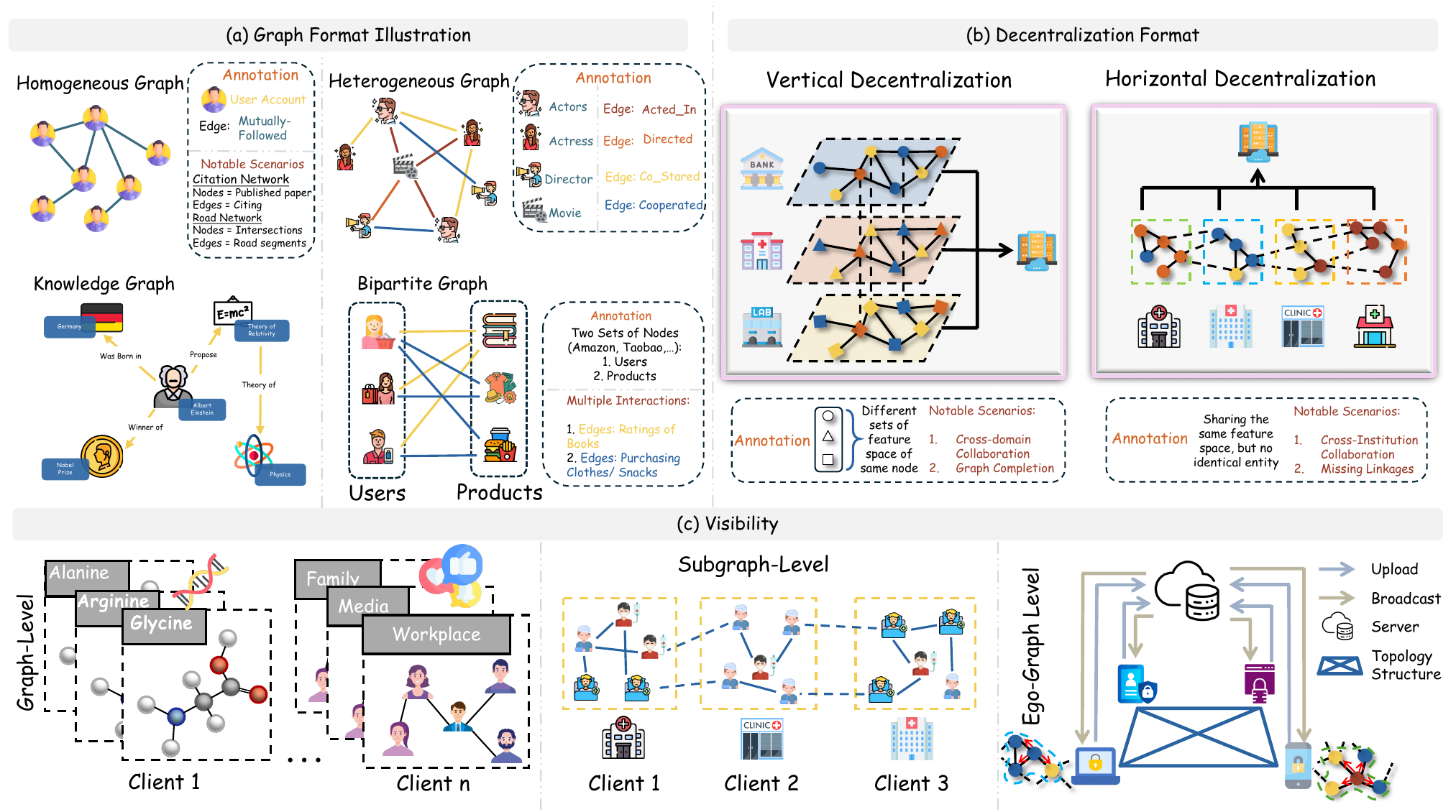}
	\caption{
        A Multi-level Illustration of three criteria of Data Characteristics Taxonomy. (a) \textbf{Graph Format Illustration} includes four graph data formats reflecting realistic scenarios with notations. (b) \textbf{Decentralization Format} displays two partition strategies, with node represent individuals in real-world scenarios. (c) \textbf{Visibility} displays three visible levels at the client-side. More details are discussed in Sec.~\ref{sec: data characteristics}.  
 }
	\label{fig: data format}
\end{figure*}
\subsection{Data Format}
\label{subsec: data format}

The data format in FGL determines the structural and semantic representation of graph data, demonstrating diversity in node types, topological structures, and feature or label information. In addition to homogeneous graphs, which are the most common representation, these criteria also encompass specialized graph types such as heterogeneous graphs, knowledge graphs, and bipartite graphs. This diversity in data formats supports specific downstream graph-related tasks and applications.

\vspace{0.1cm}
\noindent 
\textbf{Homogeneous Graph} is the simplest form of graph data, consisting of nodes and edges, some of which are associated with features. In this structure, all labeled nodes share a uniform set of attributes, and the edges connecting these nodes are semantically uniform, which indicates a consistent type of relationship between nodes. Their widespread adoption in existing literature establishes them as a fundamental building block for evaluating methods' efficiency, such as node classification~\cite{qu2023semi_node_bg}, link prediction~\cite{Zhang18link_prediction1}, and graph classification~\cite{ma2019graph_classification2}.

\vspace{0.1cm}
\noindent \textbf{Heterogeneous Graph} refers to a complex network structure characterized by multiple types of nodes and edges, each representing different entities (e.g., users, items) and relationships (e.g., user-item interactions, item-item similarities, user-user relationships). This structural diversity enables it to capture multifaceted and diverse interactions that exist in real-world scenarios, such as a healthcare network revealing correlations between patients, doctors, and prescribed drugs. Common downstream tasks on heterogeneous graphs include node classification~\cite{FedLIT}, link prediction~\cite{gu2023fedda}, and recommendation~\cite{yan2024fedhgnn}, which are more complex than their homogeneous counterparts.

\vspace{0.1cm}
\noindent \textbf{Knowledge Graph} is a specialized heterogeneous graph aimed at representing structured relationships between entities. Unlike general heterogeneous graphs, it employs predefined schemas and strict semantic constraints~\cite{KG_Basics}, modeling relationships as ontologically grounded triplets: \textit{head entity, relation, tail entity}. This structured approach enhances both machine readability and human interpretability, making it valuable for downstream tasks~\cite{KG_survey} such as relation extraction~\cite{chen2022meta_kg_sub} and knowledge graph completion~\cite{sun2024federated_kg_sub}, where the focus is on enriching semantic understanding, enabling logical reasoning, and supporting inference~\cite{KG_Reasoning_survey}. These tasks are central for two reasons: (1) extracting relations that carry explicit semantics (e.g., `authored by') from unstructured data; and (2) completing knowledge graphs for ensuring logical consistency and semantic accuracy. Together, relation extraction and knowledge completion work in tandem to enable continuous expansion and refinement of the graph. In short, the structured, interpretable, and relation-focused characteristics of knowledge graphs drive research in tasks that enhance relational understanding.

\vspace{0.1cm}
\noindent \textbf{Bipartite Graph} Nodes are divided into two distinct sets, with no nodes within a set being adjacent to one another. This graph type is commonly used in recommendation systems, depicting the interaction between users and items. These scenarios can be divided into single-user and multi-user formats, depending on the visibility structure of the graph. The former typically models each client as a user’s device~\cite{qu2023semi_node_bg}, while the latter represents each client as an e-commerce institution~\cite{li2022federated_sub_bg}.

\subsection{Decentralization Format}
\label{subsec: decentralization format}

As the second criteria under Data Characteristics, the decentralization format captures how client data is partitioned in FGL. It describes how data and learning tasks are distributed across clients, typically based on the alignment or divergence of their feature and label spaces. This aspect plays a key role in shaping the collaboration paradigm and guiding how information is exchanged across clients’ local subgraphs. We now introduce the two most commonly adopted decentralization formats: horizontal and vertical decentralization.

\vspace{0.1cm}
\noindent \textbf{Horizontal Decentralization}
describes settings in which clients share an identical feature space but hold distinct sets of labeled samples. That is, data across clients exhibit homogeneous features but non-overlapping entities. For example, in graph-based settings, each client may possess a subgraph composed of the same entity types, edge types, and features, but derived from geographically or contextually isolated regions. Collectively, these subgraphs form a global network. Such scenarios frequently occur in social network analysis, where users or organizations maintain separate subgraphs and are restricted from sharing raw data directly. Horizontal decentralization enables clients to collaboratively train a global model by exchanging model updates (e.g., gradients or embeddings) to learn both feature and structural information while maintaining data privacy. This approach is particularly well-suited for tasks such as node classification and link prediction.

\vspace{0.1cm}
\noindent \textbf{Vertical Decentralization}
describes scenarios where clients hold different types of features for the same set of entities, resulting in heterogeneous feature spaces but shared samples. This occurs in graph-based settings where clients hold complementary features for the same nodes, as seen in cross-domain collaborations (e.g., healthcare or finance) where institutions store different information—like medical records, insurance claims, or transaction histories—about the same individuals. 
An effective implementation demands sophisticated alignment techniques, such as secure entity matching and feature aggregation, to enable collaborative learning. This approach is particularly effective for tasks like graph representation learning, where aggregating complementary features improves the expressiveness of node or edge representations.

\subsection{Visibility Format}
The visibility format specifies the extent of graph information accessible to each client. It ranges from full access to the entire graph (Graph-oriented FGL), partial access to a local subgraph (Subgraph-oriented FGL), or restricted access to ego-graphs centered around individual nodes (Ego-graph-oriented FGL).

\vspace{0.1cm}
\noindent \textbf{Graph-oriented FGL} In graph-oriented FGL, each client maintains a set of topologically independent graphs, rather than a partial subgraph derived from a shared global graph. Within each client’s graph set, graphs typically share similar structures and semantic types, while domain-specific variations emerge across clients. Graph-oriented FGL seeks to identify structural patterns across domains that support effective knowledge transfer between clients. In this context, knowledge aggregation focuses on graph-level downstream tasks, such as classification or regression. Real-world applications include biomedical domains~\cite{pfeifer2023fedapp_gnn_bio1,wufederated2023fedapp_gnn_bio2} and recommendation systems~\cite{liu2022_app_social_network_1,xu2024_app_social_network_2}.

\vspace{0.1cm}
\noindent \textbf{Subgraph-oriented FGL} In subgraph-oriented FGL, each client maintains an induced subgraph derived from a shared but implicit global graph, resulting in interdependent local graphs rather than isolated partitions. Although these interconnections are not directly observable due to decentralization, the subgraphs exhibit inherent structural and statistical correlations—reflected in similarities or differences in node features, topologies, and label distributions. This setting is particularly suited for node- and edge-level tasks, including node classification, node regression, and link prediction. Realistic applications include social network analysis~\cite{hu2024_app_social_network_3}, node-level financial fraud detection~\cite{hyun2023app_gnn_fina2}, and link-level recommendation systems~\cite{user-item_recom}.

\vspace{0.1cm}
\noindent \textbf{Ego-graph-oriented FGL} In ego-graph-oriented FGL, each client accesses one or more induced k-hop ego-graphs derived from an implicit global graph. In this structure, only a central node and its surrounding neighborhood—limited to limited number of hops—are visible, along with the associated topology and features. 
This format can be considered a specialized instance of subgraph-oriented FGL, distinguished by its more restricted visibility while supporting the same downstream tasks, such as node classification and link prediction. 
Ego-graph-oriented FGL is especially suited to privacy-sensitive settings, where only localized graph information can be accessed~\cite{zhang2023fedego}. 




\subsection{Combination Summarization}
\label{subsec: data combinations}
    \begin{table*}[]
    \fontsize{8pt}{10pt}\selectfont
    \caption{FGL Studies Summarized In Data Characteristics Taxonomy}
    \centering
    \label{tab: data characteristics}
    \renewcommand{\arraystretch}{1.5}
    \begin{adjustbox}{width=1\textwidth}
    \begin{tabularx}{\linewidth}{ 
        >{\centering\arraybackslash}p{1.8cm}|
        >{\centering\arraybackslash}p{3cm}|
        >{\centering\arraybackslash}p{3.5cm}|
        >{\centering\arraybackslash}X
    }
    \toprule
    \textbf{\makecell{Data \\ Format}} & 
    \textbf{\makecell{Decentralization \\ Format}} & 
    \textbf{\makecell{Visibility \\ Format}} & 
    \textbf{Approach} \\
    
    \midrule
    
    \multirow{8}{*}{\centering Homo.} & \multirow{6}{*}{\centering Horizontal} & Graph-oriented & \makecell{GCFL\cite{xie2021gcfl}, FedStar\cite{tan2023fgl_fedstar}, FedSSP\cite{tan2024fedssp},FGCL\cite{FGCL}, FedVN~\cite{fu2024virtual}, \\
    Opt-GDBA\cite{Opt-GDBA}, NI-GDBA\cite{NI-GDBA}, FedGMark\cite{yang2024fedgmark}} \\ 
    \cline{3-4}
    & & Subgraph-oriented & \makecell{FedGraph~\cite{chen2021fedgraph}, FedGCN\cite{fedgcn}, GraphFL~\cite{wang2022graphfl},FedGL~\cite{chen2021fedgl}, FedPUB~\cite{baek2023personalized}, \\
    FedGTA\cite{li2024fedgta}, AdaFGL\cite{li2024adafgl}, FedTAD\cite{zhu2024fedtad}, FedEgo\cite{zhang2023fedego},FGGP\cite{wan2024fgl_fggp},\\
    FGSSL\cite{huangfgl_fgssl}, FedSage\cite{zhang2021fedsage}, FedDep\cite{zhang2024fgl_feddep}, FedPPN\cite{liu2024model}, FedSG\cite{wang2024fedsg}, \\FedSpray~\cite{fu2024federated}, FedCog\cite{FedCog}, HiFGL\cite{guo2024hifgl}, FedIIH\cite{FedIIH}, FedGC\cite{FedGC}}\\
    \cline{3-4}
    & & \multirow{2}{*}{Ego-Graph-oriented} & \makecell{LPGNN~\cite{LPGNN}, Fedwalk\cite{pan2022fedwalk}, Lumos\cite{pan2023lumos_node_homo}, FedSCem~\cite{FedSCem}, CNFGNN\cite{CNFGNN}}\\
    \cline{2-4}
    & Vertical & Subgraph-oriented & \makecell{GLASU\cite{zhang2023glasu}, SVFGNN\cite{liu2024svfgnn}, VFGNN\cite{VFGNN}}\\
    
    \midrule
    
    \multirow{3}{*}{Hete.} & \multirow{2}{*}{\centering Horizontal} & Subgraph-oriented & \makecell{FedAHE~\cite{FedAHE}, FedHG\cite{yan2024federated_node_hg}, FedDA~\cite{gu2023fedda}, FedHGN~\cite{FedHGN}, HGFL+~\cite{yan2024self_sub_hg}, \\
    MAFedHGL~\cite{MAFedHGL}, FedLIT\cite{FedLIT}, FedHGN\cite{fu2023fedhgn}, \\FedGNN\cite{FedGNN}, FedSoG\cite{liu2022_app_social_network_1}}\\ 
    \cline{3-4}
    & & Ego-Graph-oriented & \makecell{FedHGNN\cite{yan2024fedhgnn}}  \\
    \cline{2-4}
    & Vertical & Subgraph-oriented & \makecell{SplitGNN\cite{wu2023splitgnn}, VerFedGNN\cite{VerFedGNN}}\\ 
    
    \midrule
    
    \multirow{2}{*}{KG} & \multirow{1}{*}{\centering Horizontal} & Subgraph-oriented & \makecell{MaKEr\cite{chen2022maker}, FedEC\cite{fedec}, FedRule~\cite{FedRule}, \\FedComp\cite{wu2023fedcomp}, FedGE\cite{chen2022fedge}, GFedKG\cite{wang2024gfedkg}, FL-GMT~\cite{FL-GMT}}\\
    \cline{2-4}
    & \multirow{1}{*}{\centering Vertical} & Ego-Graph-oriented &  \makecell{Fedrkg~\cite{yao2023fedrkg_node_kg}, FedKGRec~\cite{ma2024fedkgrec_node_kg}}\\
    
    \midrule
    
    \multirow{4}{*}{Bip. Graph} & \multirow{4}{*}{\centering Horizontal} & Subgraph-oriented & \makecell{FGC\cite{yin2022fgc}, FDRS\cite{li2023fdrs_sub_bg}, FL-GMT\cite{FL-GMT}, FedFast\cite{muhammad2020fedfast}, FedRec\cite{lin2020fedrec}, \\RVEA\cite{RVEA}, FedTrans4Rec\cite{fedtrans4rec}, FedPOIRec\cite{perifanis2023fedpoirec},\\ FedNCF\cite{FedNCF}, FedSafe\cite{vyas2023federated}}\\
    \cline{3-4}
    & & \multirow{1}{*}{Ego-Graph-oriented} & \makecell{FedGNN\cite{FedGNN}, FedSoG\cite{liu2022_app_social_network_1}, PerFedRec\cite{perfedrec} , SemiDFEGL~\cite{qu2023semi_node_bg},\\ GPFedRec\cite{zhang2024gpfedrec}, PerFedRec++~\cite{luo2024perfedrec++}, HeteFedRec\cite{yuan2024hetefedrec},\\
    CDCGNNFed~\cite{CDCGNNFed}, FedGCDR\cite{yang2024federated}, CFedGR\cite{wang2024cluster}}\\
    \bottomrule
    \end{tabularx}
    \end{adjustbox}
    \end{table*}
Based on the criteria outlined above, we categorize existing FGL studies in this subsection, which provides data-centric interpretations for each category, offering insights to help researchers map these settings to their specific research objectives.
To ensure clarity, we organize the taxonomy primarily by data format. Table~\ref{tab: data characteristics} displays all categories and serves as a comprehensive guide for the following content. 

\subsubsection{\textbf{Homogeneous Graph, Horizontal, Graph-Oriented}}
In this scenario, each client owns several local graphs drawn from the same domain, so they share a common feature and label space, yet domain heterogeneity still exists across clients. Foundational work such as GCFL \cite{xie2021gcfl} shows that exchanging implicit structural knowledge among cross-domain graph datasets can yield complementary benefits. Consequently, the key challenges are, first, to mitigate the interference that arises when aggregating heterogeneous features and labels, and, second, to maximize the extraction and sharing of structural knowledge so as to boost the performance of downstream tasks, especially graph-level classification.

\subsubsection{\textbf{Homogeneous Graph, Horizontal, Subgraph-Oriented}}
\noindent
This scenario assumes that each client holds an induced subgraph of an implicit homogeneous global graph, with all clients sharing the same feature and label space. It is widely studied in FGL, as it mirrors real-world cases where multiple entities with similar data domains collaboratively train a more robust and generalizable global model. This setting directly addresses the challenge of limited data at individual clients—a core motivation for data-centric FGL. Practical applications include e-commerce companies improving product classification or recommendation systems without exposing consumers’ personal data or purchasing behavior.

\subsubsection{\textbf{Homogeneous Graph, Horizontal, Ego-Graph Oriented}}
\noindent
This configuration represents a highly constrained setting in which each client corresponds to a single node in the implicit global graph, resulting in minimal accessible training data, limited to ego-centric subgraphs that capture the immediate neighborhood of each central node. 
It is particularly well-suited to device-level applications, such as personalized recommendation systems or targeted advertising~\cite{nFedGNN}, where each user’s device maintains a local ego-graph encoding their interaction history or behavioral context. 

\subsubsection{\textbf{Homogeneous Graph,Vertical, Subgraph-Oriented}}
\noindent
In this setting, clients operate on the same set of nodes derived from a global graph but differ in the edge sets and feature spaces they locally maintain. 
As a result, each local graph can be interpreted as an edge-induced subgraph, with clients holding disjoint features for the shared nodes. 
For instance, a social media platform and a financial institution may both collect data on the same users but observe different modalities—such as social interactions versus financial transactions. 
By integrating these complementary views, the organizations can collaboratively model user profiles and preserve data privacy.

\subsubsection{\textbf{Heterogeneous Graph, Horizontal, Subgraph-Oriented}}
\noindent
Being most aligned with its homogeneous counterparts, methods adopt in this setting perform the node-level or edge-level downstream tasks on heterogeneous graphs. 
FedHG underscores the need to exploit the rich semantics embedded in \textit{meta‑paths}—relational sequences that connect heterogeneous node pairs. The main goals are to complete missing meta‑path semantics and to train heterogeneous‑graph models within a federated framework. 
A representative example includes multiple medical institutions collaboratively training graph models for disease-treatment analysis. 
Similarly, user preferences can be predicted in a heterogeneous user-item interaction graph~\cite{carlson2010toward, fu2020magnn}, where entities such as users, items, and categories form multi-typed nodes and edges.  

\subsubsection{\textbf{Heterogeneous Graph, Horizontal, Ego-Graph Oriented}} Each client maintains an ego-graph extracted from a shared heterogeneous global graph, capturing the local neighborhood of a central node along with its diverse typed relationships and attributes—while omitting inter-correlations between central nodes.
The core challenge is to enrich local data by exploring these inter-node correlations to extract more informative semantics across clients~\cite{yan2024fedhgnn}.
While feature spaces are consistent across clients, structural heterogeneity remains across each ego-graph.
This setting is well-suited for personalized applications; for example, in E-commerce platforms, a client may represent a user whose ego-graph includes interactions with products, sellers, and reviews.~\cite{yan2024federated_node_hg}
The major data-centric challenge is mitigating data heterogeneity across clients.

\subsubsection{\textbf{Heterogeneous Graph, Vertical, Subgraph-Oriented}} In this setting, each client holds the same set of nodes from a heterogeneous global graph but maintains a distinct edge set and feature space.
Unlike vertically partitioned homogeneous graphs, heterogeneous graphs encode richer semantics through diverse entity and relation types. However, incomplete labels and missing features at the client level hinder training by reducing the quality of local pattern extraction—an issue only a few studies have addressed~\cite{SplitGNN}.
For instance, specialized hospitals may each construct a subgraph of electronic health records (EHRs) for the same patient cohort, focusing on different diseases or treatment modalities. Collaborative training across such subgraphs supports advanced applications, including disease-pattern analysis, outcome prediction, and treatment recommendation~\cite{johnson2016mimic, pollard2018eicu}.

\subsubsection{\textbf{Knowledge Graph, Horizontal, Subgraph-Oriented}} 
In this setting, each client holds an induced subgraph of an implicit global knowledge graph, with aligned feature spaces across clients. Each subgraph represents a local view of the knowledge graph, containing entities and relations relevant to the client’s context. This configuration is well-suited for decentralized tasks such as entity classification, link prediction, and knowledge graph completion~\cite{chen2022meta_kg_sub}. 

\subsubsection{\textbf{Knowledge Graph, Horizontal, Ego-Graph Oriented}} Here, each client maintains an ego-graph extracted from an implicit global knowledge graph, centered around a specific entity. The ego-graph includes the entity's immediate neighbors and associated relations, while sharing a common feature space with other clients. This structure supports entity-level tasks such as personalized entity classification and recommendation~\cite{yao2023fedrkg_node_kg}.

\subsubsection{\textbf{Bipartite Graph, Horizontal, Subgraph-Oriented}} In this setting, each client holds a subgraph of an implicit global bipartite graph. A bipartite graph consists of two disjoint node sets (e.g., users and items, authors and papers), where edges only connect nodes from different sets. Each client’s subgraph includes a subset of these nodes and edges, reflecting a localized view of the global structure. This configuration is particularly applicable to decentralized tasks such as recommendation, link prediction, and community detection~\cite{li2022federated_sub_bg, li2023fdrs_sub_bg}. 

\subsubsection{\textbf{Bipartite Graph, Horizontal, Ego-Graph Oriented}} In this scenario, each client maintains an ego-graph induced from the global bipartite graph, centered on a node from one of the two disjoint sets. The ego-graph captures the node’s immediate neighbors from the opposite set, preserving the bipartite structure. This setup is suitable for node-level tasks such as personalized recommendation and link prediction in federated environments~\cite{qu2023semi_node_bg}.

\section{Data Utilization}
\label{sec: data utilization}
As the second-level taxonomy of this survey, Data Utilization reshape the technological taxonomies of previous surveys with the focus on analyzing how each FGL work addresses the data-centric challenges through three key criteria, which concern both data-centric motivations and training procedures: \\
\noindent
(1) (Sec.~\ref{subsec: utilizing position}) The positional aspect delineates at which specific stages of the FGL training procedure each study  utilizes local data, extracts knowledge from collected data, transmits the acquired knowledge proxies to the server, and applies privacy-preserving techniques to mitigate potential privacy leakage.\\
\noindent
(2) (Sec.~\ref{subsec: utilizing stage}) The sequential aspect examines in detail how each method introduces its proposed modules across different phases of the training procedure. Together with the positional aspect, this criterion provides a comprehensive understanding of how each study addresses specific data-centric challenges. Since most FGL studies follow the client-server collaborative framework, extending this classical interpretation into four distinct phases offer more nuances on each method's design.\\
\noindent
(3) (Sec.~\ref{subsec: utilizing motivation}) The third aspect, examining motivational underpinnings from a data-centric perspective, differs from the previous two aspects by explicitly highlighting the specific data-related challenges each FGL method aims to address. Placing data at the center of the discussion, this aspect details the primary data concerns targeted by each FGL method, including data quality, quantity, collaboration, efficiency, and privacy.\\
In Sec.~\ref{subsec: utilization combination}, we summarize prominent FGL studies into unique combinations of above three defined criteria.\\
Notably, several FGL studies propose innovative designs spanning multiple phases, and thereby appear across various categories as summarized in Table~\ref{tab:my_label}. This survey aims to  comprehensively reflect their underlying data-centric intentions.
Fig.~\ref{fig: data utilization} visualizes the FGL training architecture from two perspectives: the macro view highlights the core architecture, while the micro view provides detailed insights. For the latter, representative methods illustrate the Initialization and Post-aggregation phases, as these two phases are not strictly necessary to complete FGL training.

\subsection{Positional Dimensions of Data Utilization}
\label{subsec: utilizing position}

Within the FGL framework, data utilization is distributed across two distinct operational locations: the client side and the server side. Each training location engages with different data types and mechanisms, shaped by privacy concerns and task-specific objectives. This section delineates the functional and structural distinctions between these two:
\begin{figure*}[t]   
	\centering
    \setlength{\abovecaptionskip}{0.3cm}
    \setlength{\belowcaptionskip}{-0.3cm}
    \includegraphics[width=\linewidth,scale=1.00]{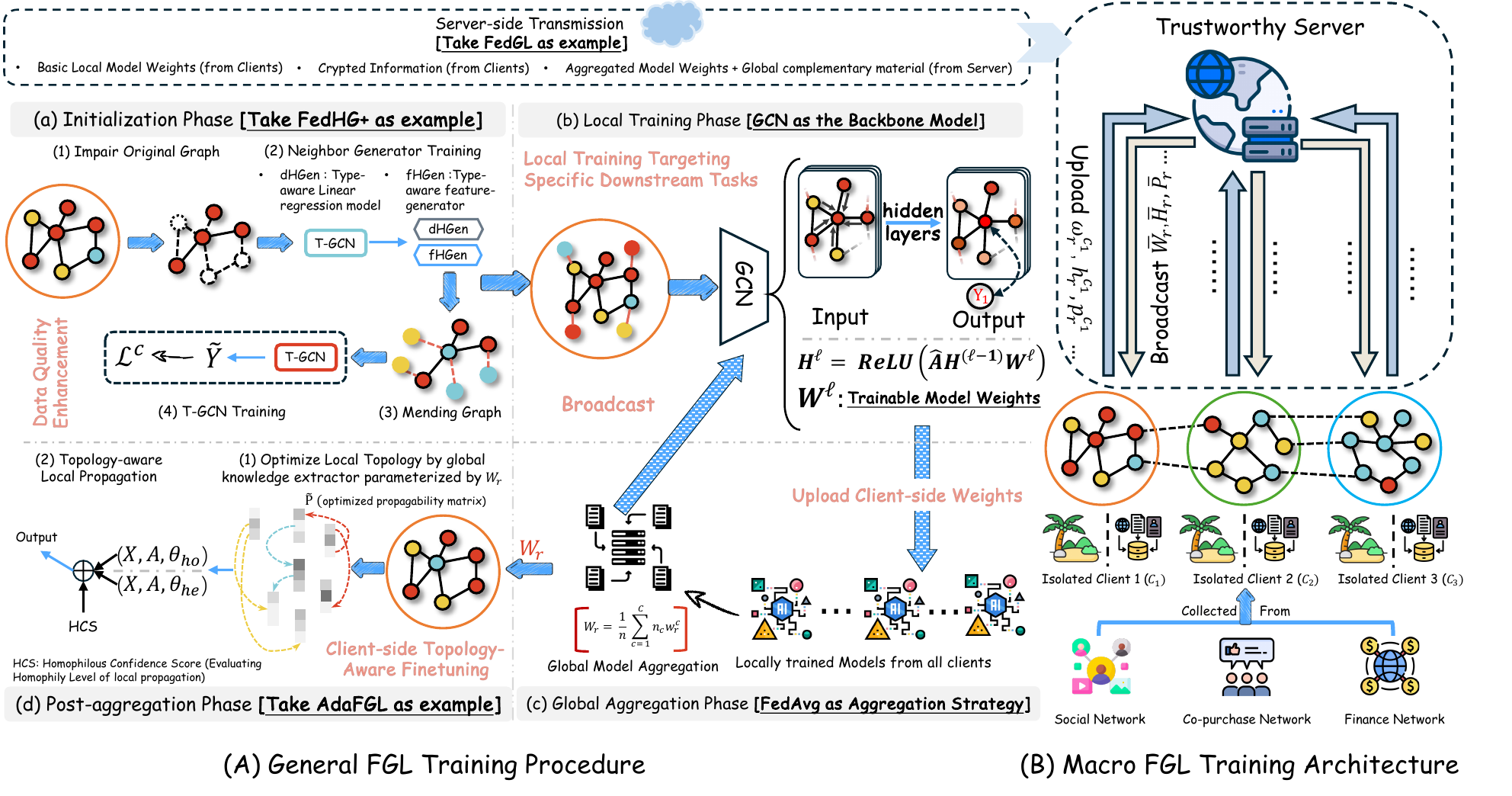}
	\caption{
       Visualizations of General FGL training architecture with representative FGL studies.
 }
	\label{fig: data utilization}
\end{figure*}

\vspace{0.1cm}
\noindent \textbf{Client-Side Data Utilization}
In FGL, each client trains a local graph model using its accessible subgraph, including local nodes, edges, and features. During training, clients transmit encrypted representations, such as node embeddings, gradients, or model parameters, to a central server for aggregation and collaborative learning. To address challenges like missing features or incomplete labels stemming from heterogeneous data, clients may incorporate auxiliary mechanisms, such as differentially private embedding sharing or feature completion strategies. Raw node-level information is never shared with the server or other clients. Instead, knowledge is communicated through privacy-preserving proxies, such as encrypted embeddings or masked gradients, that preserve essential structural and semantic patterns while protecting sensitive data.

\vspace{0.1cm}
\noindent \textbf{Server-Side Data Utilization}
The central server acts as an orchestrator, aggregating desensitized information from clients, such as model updates or desensitized embeddings, and distributing global model parameters or prototypes to guide subsequent local training. In most cases, the server does not directly access raw or client-specific data. However, depending on task design, it may leverage auxiliary proxy datasets, such as publicly available or synthetically generated graphs, to support training. These datasets are statistically independent and are often used to calibrate aggregation weights or mitigate distributional shifts across heterogeneous clients.

\subsection{Sequential Dimensions of Data Utilization}
\label{subsec: utilizing stage}

Data flow in FGL involves repeated bidirectional communication between clients and the server, with data utilization structured across four key stages: Initialization, Local Training, Global Aggregation, and Post-aggregation. Each stage targets specific challenges, balancing data efficiency with privacy and security guarantees. The following sections examine each stage and its role within the FGL framework.

\vspace{0.1cm}
\noindent \textbf{Initialization Phase}
During the initialization phase, clients perform preparatory tasks such as feature engineering that extracts and transforms graph attributes, and subgraph optimization, including topological adjustments (e.g., edge pruning, node sampling) and attribute enhancement techniques (e.g., dimensionality reduction, noise filtering). These steps improve graph quality and facilitate the subsequent FGL training. 

\vspace{0.1cm}
\noindent \textbf{Local Training Phase}
During local training, the client downloads the global model, performs local training, and uploads updated model parameters or gradients to the server. Clients may also transmit auxiliary representations, such as differentially private embeddings or local statistics, to improve global optimization without exposing raw data.

\vspace{0.1cm}
\noindent \textbf{Global Aggregation Phase}
In this phase, the server aggregates model updates, such as gradients or parameters, using federated averaging or alternative strategies that incorporate auxiliary information from clients. The updated global model is then distributed to each client for the next round of training.

\vspace{0.1cm}
\noindent \textbf{Post-aggregation Phase}
In this phase, each client can refine the global model by either leveraging domain-specific statistics from its local dataset or incorporating global knowledge to correct for local distributional bias. Localized regularization is then applied to generate personalized predictions.

\subsection{Motivational Underpinnings of Data Utilization}
\label{subsec: utilizing motivation}

In FGL, local training fundamentally depends on client-side data. In practice, FGL faces several data-related challenges, including limited data quality and quantity, inefficient collaboration, and stringent privacy constraints. Effective data utilization is essential to overcoming these challenges and improving FGL performance. This section provides a systematic analysis of these issues and the strategies proposed to address them.

\vspace{0.1cm}
\noindent \textbf{Data Quality}
Client-side data quality issues in FGL often include noise and sparsity in graph features, topology, or labels. These deficiencies can cause inaccurate representation learning and disrupt information propagation during training.

\vspace{0.1cm}
\noindent \textbf{Data Quantity}
Data quantity issues in FGL often arise on the client side, particularly in the form of label imbalance. This can lead local models to overfit majority classes while neglecting minority ones, reducing overall model generalization.

\vspace{0.1cm}
\noindent \textbf{Data Collaboration}
Collaboration-related data challenges in FGL often stem from client heterogeneity, including imbalanced sample sizes, uneven label distributions, and divergent feature spaces. These discrepancies lead to model heterogeneity, which can hinder convergence during global aggregation.

\vspace{0.1cm}
\noindent \textbf{Data Efficiency}
The quintessential data efficiency issues include excessively large graph sizes or irregular data distributions on the client side, leading to significant communication overhead between the server and clients or slow convergence during local model training. These challenges ultimately result in inefficiencies in the overall FGL process.

\vspace{0.1cm}
\noindent \textbf{Data Privacy}
Data privacy concerns in FGL stem from the sensitivity of client-held data, which poses risks of information leakage during communication with the server, and enhances the robustness of FGL methods under malicious attacks~\cite{FedTGE_ICLR25}.
\begin{center}
    \begin{table*}[ht]
    \fontsize{8pt}{10pt}\selectfont
    \caption{FGL Studies Summarized In Data Utilization Taxonomy}
    \centering
    \renewcommand{\arraystretch}{1.5}
    \begin{adjustbox}{width=1\textwidth}
    \begin{tabularx}{\linewidth}{ 
        >{\centering\arraybackslash}m{2.8cm}|
        >{\centering\arraybackslash}m{2.5cm}|
        >{\centering\arraybackslash}m{3cm}|
        >{\centering\arraybackslash}X
    }
    \toprule
    \textbf{\makecell{Motivational \\ Underpinnings}} & 
    \textbf{\makecell{Positional \\ Dimensions}} & 
    \textbf{\makecell{Sequential \\ Dimensions}} & 
    \textbf{Approach} \\ 
    \midrule
    \multirow{2}{*}{\makecell{Data \\ Quality}} & \multirow{2}{*}{Client-side} & Initialization& \makecell{FedHG+~\cite{zhang2022subgraph}, HGFL~\cite{yan2024self_sub_hg}, FedHGNN~\cite{yan2024federated_node_hg}, SemiDFEGL~\cite{qu2023semi_node_bg},\\ FedSage~\cite{zhang2021fedsage}, FedDep~\cite{zhang2024fgl_feddep}, FedCog~\cite{FedCog}}\\
    \cline{3-4}
    & & Post-aggregation &\makecell{AdaFGL~\cite{li2024adafgl}}\\
    \midrule
    \makecell{Data \\ Quantity} & {Client-side} & Local Training& \makecell{FedSpray~\cite{fu2024federated} , OpFGL~\cite{yan2024towards}, FedLoG~\cite{kim2025subgraph}, \\FedKGRec~\cite{ma2024fedkgrec_node_kg}, FDRS ~\cite{li2023fdrs_sub_bg}, Lumos ~\cite{pan2023lumos_node_homo}, HeteFedRec\cite{yuan2024hetefedrec}}\\
    \midrule
    \multirow{5}{*}{\makecell{Data \\ Collaboration}} & \multirow{2}{*}{Client-side} & Initialization& \makecell{FedTAD~\cite{zhu2024fedtad}, GraphGL~\cite{wang2022graphfl}, FedLIT~\cite{FedLIT}, HGFL+~\cite{yan2024self_sub_hg}}\\ 
    \cline{3-4}
    & & Local Training & \makecell{FedGTA~\cite{li2024fedgta}, FGSSL~\cite{huangfgl_fgssl}, FGGP~\cite{wan2024fgl_fggp}, FedStar~\cite{tan2023fgl_fedstar}, FedVN~\cite{fu2024virtual}\\
    VerFedGNN~\cite{VerFedGNN}, MaKEr~\cite{chen2022maker}, FedTrans4Rec~\cite{fedtrans4rec}, PerFedRec~\cite{perfedrec}}\\
    \cline{2-4}
    & \multirow{3}{*}{Global-side} & \multirow{3}{*}{Global Aggregation} & \makecell{FedGL~\cite{chen2021fedgl}, GCFL~\cite{xie2021gcfl}, FedEgo~\cite{zhang2023fedego}, FedSSP~\cite{tan2024fedssp} , FedPUB~\cite{baek2023personalized}, \\ FGC\cite{yin2022fgc},  GPFedRec\cite{zhang2024gpfedrec}, CDCGNNFed\cite{CDCGNNFed}, FedComp\cite{wu2023fedcomp}, \\ FedSG~\cite{wang2024fedsg}, FedIIH~\cite{FedIIH}, FedGC~\cite{FedGC}, CNFGNN~\cite{CNFGNN}, MAFedHGL~\cite{MAFedHGL},\\ 
    FedEC~\cite{fedec}, FedGE~\cite{chen2022fedge}, GFedKG~\cite{wang2024gfedkg}, FedNCF~\cite{FedNCF}, PerFedRec~\cite{perfedrec}} \\
    \midrule
    \multirow{3.5}{*}{\makecell{Data \\ Efficiency}} 
    & \multirow{2}{*}{Client-side} & \multirow{2}{*}{Local Training} & \makecell{Fedgraph~\cite{chen2021fedgraph}, FedDA~\cite{gu2023fedda}, FedSCem ~\cite{FedSCem}, FedGCN~\cite{fedgcn} \\FedHG+~\cite{fedhg}, Lumos\cite{pan2023lumos_node_homo}, FedWalk\cite{pan2022fedwalk}, LPGNN~\cite{LPGNN}, SVFGNN~\cite{liu2024svfgnn}}\\
    \cline{2-4}
    & Global-side & Global Aggregation &\makecell{FedAHE~\cite{FedAHE}, PerFedRec++~\cite{luo2024perfedrec++},  CDCGNNFed~\cite{CDCGNNFed}, Lumos\cite{pan2023lumos_node_homo}, \\ GLASU~\cite{zhang2023glasu}, FedSCem~\cite{FedSCem}, FedWalk\cite{pan2022fedwalk}, FedFast~\cite{muhammad2020fedfast}}\\
    \midrule
    \multirow{5.5}{*}{\makecell{Data \\ Privacy}}
    & \multirow{3.5}{*}{Client-side} & \multirow{3.5}{*}{Local Training} & \makecell{FedHGN~\cite{fedhg}, LPGNN~\cite{LPGNN}, FedSCem~\cite{FedSCem},FedGNN~\cite{FedGNN}, \\FeSoG~\cite{liu2022federated}, Opt-GDBA~\cite{Opt-GDBA}, NI-GDBA~\cite{NI-GDBA} 
    FGCL~\cite{FGCL} , \\ Lumos\cite{pan2023lumos_node_homo}, FedLIT~\cite{FedLIT}, FedWalk\cite{pan2022fedwalk}, FedGCDR\cite{yang2024federated}\\
    nFedGNN\cite{nFedGNN},FedDep~\cite{zhang2024fgl_feddep},HiFGL~\cite{guo2024hifgl}, FedGC~\cite{FedGC},\\ SVFGNN~\cite{liu2024svfgnn}, FedGNN~\cite{FedGNN}}\\
    \cline{2-4}
    & Global-side & Global Training & \makecell{Lumos\cite{pan2023lumos_node_homo}, FedSCem~\cite{FedSCem}, FedHGN~\cite{fedhg}, FedLIT~\cite{FedLIT},\\
    FedWalk\cite{pan2022fedwalk}, FedGMark~\cite{yang2024fedgmark}, nFedGNN\cite{nFedGNN}, FedGNN\cite{FedGNN},\\  FedRule~\cite{FedRule},VFGNN~\cite{VFGNN}, VerFedGNN~\cite{VerFedGNN}, FedRec~\cite{lin2020fedrec},\\ RVEA~\cite{RVEA}, FedPOIRec~\cite{perifanis2023fedpoirec}}\\
    \bottomrule
    \end{tabularx}
    \end{adjustbox}
    \label{tab:my_label}
    \end{table*}
\end{center}

\vspace{-1.3cm}
\subsection{Combination Summarization}
\label{subsec: utilization combination}
\subsubsection{\textbf{Data Quality, Client-side, Initialization Phase}} Before local training, methods in this category preprocess data to mitigate client heterogeneity by enriching local graph information.
FedHG+~\cite{fedhg} employs federate-trained generators to synthesize missing neighbor entities, aligning each client's subgraph distribution closer to the global graph.
HGFL~\cite{yan2024self_sub_hg} uses a generator-based method to reconstruct the adjacency matrix among existing nodes, enhancing subgraph connectivity without introducing artificial entities.
FedHGNN~\cite{yan2024federated_node_hg} applies Semantic-Preserving User–Item Interaction Publishing, which shares distilled user-item interactions among FL clients in a preliminary disclosure to reduce local bias.
SemiDFEGL~\cite{qu2023semi_node_bg} creates virtual public items by leveraging server-provided auxiliary information, indirectly connecting heterogeneous subgraphs across clients.
These preprocessing techniques improve data consistency and representation, positively impacting subsequent federated graph learning performance.

We illustrate the Graph-mending technique of HGFL~\cite{yan2024self_sub_hg} as an example, which does not invite new generative nodes but rely on existing nodes of the original heterogeneous graph to reconstruct the adjacency matrix of each meta-path, which carries vital semantic information in heterogeneous graphs. 
\begin{equation}
  h_i^{P_n}=\frac{A_{ii}}{d_i+1}x_i+\sum_{j\in N_i^{P_n}}\frac{A_{ij}}{\sqrt{(d_i+1)(d_j+1)}}x_j,  
\end{equation}
\begin{equation}
\label{eq: HGFL_progability}
E_{ij}^{P_n}=\text{Softmax}(\sigma(h_i^{P_n}\cdot S\cdot h_j^{P_n})),
\end{equation}
\begin{equation}
A_{mend}^{P_n} = E^{P_n} + A^{P_n},
\end{equation}
where $h_i^{P_n}$ is the embedding representation output of any GNN-based encoder of private node $i$, under the meta-path $p_n$. and $j\in N^{p_n}_i$ is the adjacent node of node $i$ within the meta-path. 

The equation~\ref{eq: HGFL_progability} delineates the probability prediction of connectivity between two nodes to recapture their correlations. $E_{ij}^{P_n}$ is then trained based on the minimizing the loss compared to the original adjacency matrix.  In the end, mending the original graph by concatenating $E_{ij}^{P_n}$, the reconstructed adjacency matrix, with original adjacency matrix of the meta-path. 

\subsubsection{\textbf{Data Quantity, Client-side, Local Training Phase}} Local graph data collected by clients are frequently imbalanced.
FedSpray~\cite{fu2024federated} alternately trains a class-wise structural proxy that regularizes client-side updates.
O-pFGL~\cite{yan2024towards} computes node-adaptive coefficients from pseudo-labels and homophily, balancing local contributions with global knowledge.
FedLoG~\cite{kim2025subgraph} synthesizes virtual samples from high-degree nodes and shares them across clients to equalize class distributions.
FedKGRec~\cite{ma2024fedkgrec_node_kg} integrates a KG-aware recommendation module that enriches sparse user–item interactions while safeguarding privacy via differential privacy.
FDRS~\cite{li2023fdrs_sub_bg} augments the bipartite graph with latent user–user and item–item edges extracted from social networks and purchase histories.
FedHGL~\cite{FedHGL} leverages global prototypes for “semantic compensation,” rectifying local biases and interpolating virtual minority nodes.
Lumos~\cite{pan2023lumos_node_homo} builds a heterogeneity-aware tree by attaching virtual parent nodes, thereby increasing representational capacity.

FedSpray tackles the imbalaced data quantity issues with its acquisition of reliable class-wise structural proxy through its fededrated trainings. Its main procedure is defined as:

\begin{small}
\begin{equation}
\mathbf{p}_{i}^{(k)}=g\left(\mathbf{x}_{i}^{(k)}, \mathbf{s}_{i}^{(k)} ; \omega\right)=g_{c}\left(\operatorname{Combine}\left(\mathbf{e}_{i}^{(k)}, \mathbf{s}_{i}^{(k)}\right) ; \omega_{c}\right),
\end{equation}
\end{small}
where $\mathbf{p}_{i}^{(k)}$ denotes the unbiased node representation produced by the feature–structure encoder $g$ parameterized by $\omega$.  
The structural proxy $\mathbf{s}_{i}^{(k)}$ is formed from the 1-hop neighbors of node $i$ that share its class label, while $\mathbf{e}_{i}^{(k)}$ is the corresponding low-dimensional feature embedding.  
Concatenating $\mathbf{e}_{i}^{(k)}$ with $\mathbf{s}_{i}^{(k)}$ yields the input to the classifier $g_{c}$.  
This coupling produces bias-reduced node embeddings—particularly for minority classes—which are further refined via distillation-based loss functions during local training.

\subsubsection{\textbf{Data Collaboration, Client-side, Initialization Phase}} To facilitate collaboration among  clients with heterogeneous data, each client can pre-process its local graph and pre-compute relevant statistics before engaging in the federated procedure. 
GraphGL~\cite{wang2022graphfl} employs model-agnostic meta-learning prior to federated training, providing a stronger initialization for adapting to each client’s data. 
FedLIT~\cite{FedLIT} clusters relation embeddings to generate local relation-type prototypes, which are uploaded to the server to guide aggregation.
FedTAD~\cite{zhu2024fedtad} computes the class-wise knowledge-reliability score for every class $c$ on each client as:
\begin{equation}
\Phi_c = \sum_{v_i \in \mathcal{V}_c^{\text{labeled}}}
         \frac{\sum_{v_j \in \mathcal{N}(v_i)}
               s\!\bigl(\mathbf{h}_i, \mathbf{h}_j\bigr)}
              {|\mathcal{N}(v_i)|},
\end{equation}
\begin{equation}
\begin{aligned}
\mathbf{h}_i &= \mathbf{x}_i \| \mathbf{h}_i^{\text{topo}},\;
\mathbf{h}_i^{\text{topo}}= [\mathbf{T}_{ii}, \mathbf{T}_{ii}^2, \dots, \mathbf{T}_{ii}^p] \in \mathbb{R}^p,
\end{aligned}
\end{equation}
where $s(\cdot, \cdot)$ denotes the cosine similarity, $\mathcal{V}_c^{\text{labeled}}$ is the labeled node set of class $c$ and $\mathcal{N}_(v_i)$ is the 1-hop neighborhood of node $v_i$. The resulting score is transmitted to the server so that clients with higher class-specific confidence exert a proportionally greater influence on the global model. 

\subsubsection{\textbf{Data Collaboration, Client-side, Local Training Phase}}
\begin{spacing}{0.9}
To enable collaboration among clients with heterogeneous data, several client-side training algorithms have been proposed. 
At the \textit{subgraph level}, FedGTA~\cite{li2024fedgta} combines label propagation with online estimates of local prediction smoothness and the mixed moments of neighbor features; these statistics are then transmitted to the server for topology-aware personalized aggregation. 
FGSSL~\cite{huangfgl_fgssl} employs contrastive learning on augmented graph views in tandem with model training, while FGGP~\cite{wan2024fgl_fggp} builds a semantically and structurally informed global view of the local graph and performs prototype-based contrastive learning throughout training.
At the \textit{graph level}, FedVN~\cite{fu2024virtual} augments each client graph with learnable virtual nodes and trains a personalized edge generator to connect them to the local graph, augmenting client-side data and mitigating distribution shifts.
FedStar~\cite{tan2023fgl_fedstar} alternately trains a feature–structure-decoupled GNN that separates attributive and structural knowledge into two parallel channels, sharing only the structure encoders to disseminate structural information. 
It construct the structure embedding that incorporate both local and global structural patterns as:
\end{spacing}
\begin{footnotesize}
\begin{equation}
s_v^{\mathrm{DSE}}
  = \bigl[\,\mathbb{I}(d_v = 1),\;
           \mathbb{I}(d_v = 2),\;
           \dots,\;
           \mathbb{I}(d_v \ge k_1)\,\bigr]
  \in \mathbb{R}^{k_1},
\end{equation}
\begin{equation}
\begin{aligned}
s_v^{\mathrm{RWSE}} &= \left[\,\mathbf{T}_{ii},\,\mathbf{T}_{ii}^2,\,\dots,\,\mathbf{T}_{ii}^{k_2}\,\right] \in \mathbb{R}^{k_2},
\;
s_v = s_v^{\mathrm{DSE}} \| s_v^{\mathrm{RWSE}},
\end{aligned}
\end{equation}
\end{footnotesize}
\noindent
where $d_v$ is the degree of node $v$, $\mathbf{T} = \mathbf{A}\mathbf{D}^{-1}$ is the random-walk transition matrix, and $k_1$ and $k_2$ denote the dimensions of DSE and RWSE, respectively.
On the client side, FedStar utilizes degree-based structural embeddings (DSE), which encode local structural information via node degrees. Because node degree is a fundamental geometric property, it can be extracted easily and efficiently from graph datasets across domains.
FedStar also adopts random-walk–based structural embeddings (RWSE) to capture domain-agnostic structural information from a global perspective.
The final structural embedding is obtained by concatenating both.

\subsubsection{\textbf{Data Collaboration, Global-side, Global Aggregation Phase}} To facilitate collaboration among clients with heterogeneous data, FedTAD~\cite{zhu2024fedtad} performs data-free distillation, transferring reliable knowledge from an ensemble of client models to the aggregated model.
FedGL~\cite{chen2021fedgl} constructs a global pseudo-graph and pseudo-labels, distributing them to clients for self-supervised learning that mitigates distribution shift.
FedGTA~\cite{li2024fedgta} derives personalized aggregation coefficients from the uploaded local prediction smoothness and mixed moments of neighbor features.
GCFL~\cite{xie2021gcfl} clusters clients by their uploaded gradient series and aggregates models cluster-wise on the server.
FedEgo~\cite{zhang2023fedego} aggregates uploaded ego-graphs to train a global GraphSAGE classifier centrally.

FedGL aggregate the uploaded client models $W_k$, node embeddings $H_k$ and client perditions $P_k$ at the server by weighted average:
\begin{equation}
\begin{aligned}
\overline{P} &= \sum_{k=1}^{K} \frac{N_k}{M} P_k,
\overline{H} = \sum_{k=1}^{K} \frac{N_k}{M} H_k,
\overline{W} = \sum_{k=1}^{K} \frac{N_k}{M} W_k,
\end{aligned}
\end{equation}
where $N_k$ is the number of nodes on client $k$ and $M$ is the sum of the number of nodes of all clients. The server generates global pseudo‑labels and assembles a pseudo‑graph reflecting the overall distribution. It then transmit them alongside the aggregated global model to clients before the next round, supplying complementary label and structural information to mitigate data heterogeneity across clients.

As a signature work of Personalized FGL,  FedPUB~\cite{baek2022fedpub} proposes to measure the functional similarity of uploaded GNNs by feeding the generated random graphs to all clients' models and then calculates  similarity of their outputs in the server. 
$\tilde{\mathbf{h}}_i$ is the averaged output of all node embeddings learned through generated random graph from client $i$'s model. 
The personalized model aggregation process is:
\begin{equation}
\begin{aligned}
\overline{\mathbf{\theta}}_i &= \sum_{j=1}^{K} \alpha_{ij}\mathbf{\theta}_j, \;
\alpha_{ij}= \frac{\exp\left(\tau \cdot \cos(\tilde{\mathbf{h}}_i, \tilde{\mathbf{h}}_j)\right)}{\sum_{k=1}^{K} \exp\left(\tau \cdot \cos(\tilde{\mathbf{h}}_i, \tilde{\mathbf{h}}_k)\right)},
\end{aligned}
\end{equation}
where $\tau$ is a hyper-parameter to scale the similarity score. By this design, FedPUB manages to share knowledge between clients with similar local subgraphs, and thus avoids the conflict of incompatible local knowledge.

\subsubsection{\textbf{Data Efficiency, Client-side, Local Training Phase}} To improve \textit{training efficiency}, FedGraph~\cite{chen2021fedgraph} employs reinforcement learning to sample informative subgraphs, thereby reducing computational overhead.
For \textit{communication efficiency}, FedPUB~\cite{baek2022fedpub} applies personalized parameter masking, and O-pFGL~\cite{yan2024towards} uploads only class-wise distribution statistics, both of which markedly lower transmission volume compared with full-parameter exchange.
FedDA~\cite{gu2023fedda} further trims communication by using Restarting-based and Exploration-based strategies that upload parameters from only a subset of clients.
To accelerate \textit{local convergence}, FedSCem~\cite{FedSCem} shares partial node embeddings across clients and aligns them via contrastive learning, while FedHG+~\cite{fedhg} introduces a sample-based normalization term during gradient computation that re-weights node contributions based on local sample sizes.

\subsubsection{\textbf{Data Efficiency, Global-side, Global Aggregation Phase}} In the server-side aggregation process, adjusting aggregation weights or carefully selecting parameters can accelerate convergence of the global model.
FedAHE~\cite{FedAHE} employs “Dynamic-weight Aggregation,” dynamically assigning weights according to the difference between each client’s version number and the global parameter version.
PerFedRec++\cite{luo2024perfedrec++} inserts an additional initialization stage that leverages privacy-preserving auxiliary signals—such as client-selection history, pseudo-item labels, and differentially private statistics—to generate multi-view representations for contrastive learning, thereby improving convergence efficiency.
CDCGNNFed\cite{CDCGNNFed} first repairs the server-side graph by predicting missing links and then trains a global knowledge model, a strategy that both corrects client-induced biases and enhances computational efficiency.

\subsubsection{\textbf{Data Privacy, Client-side, Local Training Phase}}
Data privacy remains a critical concern in FGL. For instance, in recommendation systems the locally learned item representations can unintentionally reveal user traits. Beyond the common defences of local differential privacy (LDP) and secure aggregation, recent research has proposed several complementary safeguards.
FedHGN~\cite{FedHGN} applies schema-weight decoupling, decomposing each relation weight into a schema-specific coefficient and a base weight; only these components are uploaded, and the schema coefficients received from other clients are heuristically aligned, thereby reducing relational privacy leakage.
LPGNN~\cite{LPGNN} encrypts local features with a modified 1-bit encoder and a robust training framework, improving inference accuracy under noisy labels while using propagation-drop for label denoising.
FedSCem~\cite{FedSCem} protects individual node embeddings through a two-stage protocol of secure alignment followed by secure aggregation.
FedGNN~\cite{FedGNN} obscures true user–item interactions by adding randomly sampled item-embedding gradients to the uploaded gradients for interacted items.
FeSoG~\cite{FedSog} injects pseudo-item interactions, which introduce noise for privacy and supply additional ratings that mitigate the cold-start problem.

\subsubsection{\textbf{Data Privacy, Global-side, Global Aggregation Phase}}
In global aggregation, naive weighted parameter averaging can expose private information.
FedHGN \cite{FedHGN} avoids such leakage by performing basis decomposition locally and uploading only the basis weights, thereby obscuring each client’s relation types.
FedLIT \cite{FedLIT} aggregates models by matching privacy-preserving relation prototypes, which implicitly fuses knowledge and avoid direct exposes of raw information.
FedRule \cite{FedRule} proposes a federated rule-recommendation framework that employs graph neural networks (GNNs) to learn personalized smart-home automation rules while preserving user privacy. Unlike traditional matrix-based recommender systems, FedRule builds a local entity graph for each user and frames recommendation as a link-prediction task. The GNN learns rule representations from the user’s own entity connections, so no raw interaction data ever leave the client device, providing privacy guarantees.


\section{Euclidean-oriented FGL} 
\label{sec: euclidean-oriented FGL}
Euclidean-oriented FGL generalizes traditional FGL by supporting non-graph data types, such as images, text, or time-series data. 
While the communication topology among clients and the server remains graph-structured, the underlying data itself is not necessarily organized as a graph. 
Unlike graph-specific variants (e.g., graph-oriented, subgraph-oriented, ego-oriented), Euclidean-oriented FGL offers greater flexibility by enabling the integration of diverse data types while retaining the core benefits of federated learning. 
It distinguishes between two forms of topology as listed below.  

\subsection{Physical Communication Topology}
Physical communication topology refers to the network pathways used for client-server or peer-to-peer communication in FGL systems. 
It is typically static and is commonly represented by structures such as the star topology~\cite{DSGD-FL,dFedU}. 
Physical communication topologies are leveraged to improve efficiency through partial and hierarchical model aggregation, enhance privacy by avoiding direct transmission of local models to a central server, and improve scalability through modular, replicable network designs. 
These topologies are generally categorized as centralized (e.g., star), decentralized (e.g., mesh), or hybrid structures that combine multiple standard designs.

\subsection{Logical Topology}
In contrast to physical topology, logical topology is a task-dependent virtual structure dynamically defined by the server based on inter-client relationships, supporting coordinated optimization in FGL. 
For example, in image classification, clients typically hold independent image samples, resulting in sparse or disconnected logical topologies. 
In traffic flow prediction, data is structured as spatiotemporal sequences, and the server constructs a denser logical topology to capture regional dependencies. 
In both cases, the server builds an adjacency matrix based on inter-client similarity~\cite{SFL,xing2022_app_social_network_4}, often derived from data distributions or feature embeddings. 
In power allocation, clients provide wireless network data, and the logical topology reflects estimated interference relationships. 
For indoor localization, clients hold Wi-Fi signal strength measurements structured as local graphs, while the server constructs a client relation graph using uploaded embeddings or parameters to capture similarity among data distributions.

\section{Data Application}
\label{sec: Data Application}
FGL has emerged as an effective framework in domains where privacy preservation, personalization, and collaborative learning are critical. To demonstrate its practical relevance, we provide detailed discussions of FGL applications across six key domains, highlighting representative studies tailored to domain-specific challenges: 

\noindent 
\subsection{Molecular Networks}
In molecular science, data is often represented as molecular structures, chemical properties, and biological interactions, which are naturally modeled as graphs or networks. FGL aids collaborative modeling of these molecular graphs by enabling clients to train together without sharing raw data—crucial for applications like drug discovery and biomedical research.
Recent advancements in \textit{Personalized Drug Discovery} focus on adapting models to dataset-specific characteristics while respecting privacy constraints. For instance, MolCFL~\cite{guo2024_app_molecular_network_2} uses generative clustered learning for client-specific personalization, while GFedKRL~\cite{NingWLYL23_app_molecular_network_6} reconstructs molecular knowledge graphs with secure communication protocols. Additionally, SpreadGNN~\cite{he2022_app_molecular_network_4} enables multi-task learning across decentralized datasets, and \cite{ZhuLW22_app_molecular_network_5} and \cite{zhu2022_app_molecular_network_8} propose feature alignment methods to address molecular data heterogeneity.
For \textit{Collaborative Disease Research}, hybrid approaches combine computational efficiency with expert insights. Ensemble-GNN~\cite{pfeifer2023_app_molecular_network_3} integrates federated ensemble learning with GNNs to identify disease modules across distributed datasets, which is further extended by \cite{hausleitner2024_app_molecular_network_1}’s human-in-the-loop framework, incorporating domain expert feedback for adaptive model weighting.
Architectural innovations also enhance FGL’s role in molecular science. Foundational work by \cite{pei2021_app_molecular_network_7} introduced decentralized FGL, while FedGCN~\cite{hu2022_app_molecular_network_11} extends FGL to non-Euclidean molecular data. Additionally, \cite{lin2020_app_molecular_network_9} offers privacy-aware optimization techniques for modeling molecular relationships.

\noindent 
\subsection{Social Networks}
Social networks capture complex user interactions, powering applications across various domains. FGL’s decentralized approach allows organizations and platforms to collaboratively learn from these interactions without sharing raw user data.
\cite{liu2022_app_social_network_1} laid the groundwork for modeling cross-platform user-item interactions without data exchange, further extended by \cite{xu2024_app_social_network_2}, which enhances the model with graph attention mechanisms to capture localized social influence and improve recommendation quality.
In \textit{Friendship Prediction}, \cite{hu2024_app_social_network_3} proposes a hybrid method combining graph convolutional autoencoders and factorization machines in a federated setting, allowing for latent connection inference without sharing adjacency data.
\textit{Big-Fed}~\cite{xing2022_app_social_network_4} introduces a bilevel optimization approach that integrates global graph topology and local client models, improving knowledge exploration while reducing communication overhead in large-scale networks.
Efficient inference is another key innovation. FedGraph-KD~\cite{wang2023_app_social_network_5} leverages knowledge distillation to compress complex graph models into lightweight versions, enabling faster, scalable inference.

\noindent 
\subsection{Recommendation Systems}
Graph datasets in recommendation systems represent user-item interactions, with nodes for users and items, and edges reflecting preferences or ratings. FGL addresses key challenges in this area with innovations focused on data protection and model performance.
Foundational works such as FedGNN~\cite{FedGNN} and FedPerGNN~\cite{FedPerGNN} incorporate differential privacy into graph-based recommendation systems within FGL. FedGR~\cite{ma2023_app_recommender_system_2} enhances privacy further with graph masking techniques.
\cite{TianXCLZ24_app_recommender_system_4} introduces an attention-based graph alignment mechanism for cross-domain knowledge transfer, while \cite{VerFedGNN} presents a secure embedding aggregation protocol for vertical FGL, enabling feature collaboration across organizations without exposing raw data.
For \textit{Personalized Recommendations}, \cite{HanHLQTG24_app_recommender_system_5} develops a meta-learning-enhanced subgraph FGL framework, adapting to local graph patterns for fine-grained personalization while maintaining local data privacy.

\noindent 
\subsection{Financial Networks}
Financial transaction graphs capture exchanges between individuals and organizations, revealing patterns in credit flow, fraud, and financial risk. FGL enables collaborative analysis for tasks like fraud detection, anti-money laundering, and credit scoring while keeping sensitive data decentralized.
\cite{suzumura2019_app_financial_network_1} pioneered FGL applications for cross-institutional financial crime detection using advanced graph pattern mining, later extended by \cite{tang2024_app_financial_network_2}, which develops GNN-based credit card fraud detection models.
\cite{tang2024_app_financial_network_4} introduces a differential privacy-preserving GNN framework for supply chain finance, encoding sensitive transaction data into privacy-protected embeddings before training. This is complemented by FedEE~\cite{xie2022_app_financial_network_5}, which enhances inter-enterprise collaboration via advanced graph-based feature extraction.
For \textit{Regulatory Compliance}, \cite{lee2022_app_financial_network_6} proposes a graph-regularized federated linear regression model for anti-money laundering tasks, incorporating a one-shot data mixing technique to ensure compliance with FINRA regulations while safeguarding financial data.

\noindent 
\subsection{Smart Cities}
Smart-city data, derived from traffic networks, energy grids, and public services, is inherently graph-structured, with nodes representing infrastructure elements and edges capturing interactions. FGL enables secure, decentralized analysis of these graphs, facilitating coordinated intelligence across urban domains.
In \textit{Adaptive Traffic Systems}, FedSTN~\cite{yuan2022_app_smart_city_3} uses spatial-temporal GNNs for decentralized traffic analysis, meeting real-time traffic management needs. \cite{liu2023_app_smart_city_2} extends this with a hierarchical federated approach for city-wide coordination, addressing data heterogeneity across districts.
For \textit{Energy Infrastructure}, \cite{li2024federated_app_smart_city_4} develops a robust GNN architecture that adapts to local energy usage and defends against data poisoning. \cite{you2024_app_smart_city_5} enhances energy management by combining meta-learning with graph convolution to transfer knowledge across urban charging networks.
In \textit{Distributed Urban Sensing}, \cite{wu2022_app_smart_city_8} advances indoor localization with federated graph attention, while \cite{jiang2022_app_smart_city_9} introduces a privacy-compliant surveillance system using dynamic GNNs to process distributed camera network data, reducing risks to citizen privacy.
Methodological innovations include cross-domain graph feature alignment~\cite{meng2021_app_smart_city_7}, secure aggregation for spatial-temporal graphs~\cite{zhang2021_app_smart_city_10}, and an edge-cloud model framework for urban computing~\cite{djenouri2023_app_smart_city_1}.

\noindent 
\subsection{Internet of Things (IoT)}
IoT ecosystems generate graph-structured data from edge devices such as sensors, appliances, and wearables, where nodes represent devices and edges capture their interactions. This data is inherently heterogeneous and non-IID, with each device producing unique, environment-specific readings that must remain on-device. As a result, computation and storage are decentralized under bandwidth and energy constraints, with sparse updates requiring efficient and robust aggregation.
In \textit{Security Threat Detection}, \cite{XingHDSHW24_app_IOT_3} develops a representation learning framework to analyze device interaction patterns while protecting sensitive data. \cite{zhang2024_app_IOT_1} introduces graph sparsification techniques to improve security during federated training, enhancing resistance to adversarial attacks~\cite{cai2024fgad, cai2024lgfgad,FedTGE_ICLR25}.
FedHGL~\cite{WeiCZHZHW24_app_IOT_4} tackles device heterogeneity by using diverse GNN architectures to bridge various modalities and communication protocols in cross-institutional federated learning settings.
For \textit{Trustworthy Edge Intelligence}, \cite{abualkishik2023_app_IOT_2} proposes a verifiable FGL system with decentralized trust mechanisms to ensure model integrity and robustness are not compromised in malicious wireless IoT environments.

\noindent 
\subsection{Health Networks}
Health networks generate diverse data, including medical images, sensor-collected physiological signals, and networked patient records, which support disease prediction and epidemic monitoring. FGL has driven three major health network areas:
\subsubsection{\textbf{Medical Imaging and Clinical Diagnosis}}
Building on \cite{bayram2021_app_health_network_1}’s brain-network analysis, FedGraphMRI-net~\cite{ahmed2025_app_health_network_18} enables cross-institutional MRI reconstruction. In mental health, \cite{hu2024_app_health_network_16} improves diagnostic accuracy with graph-convolutional aggregation, while \cite{ahmed2022_app_health_network_5} enhances symptom detection using hypergraph attention mechanisms.
\subsubsection{\textbf{Electronic Health Records (EHR) Analysis}}
Early work by \cite{brisimi2018_app_health_network_13} is expanded by FedNI~\cite{peng2022_app_health_network_9}, which introduces network inpainting for population health prediction. Recent efforts include personalized methods for non-IID EHR data~\cite{tang2024_app_health_network_6}, and COVID-19 forecasting models~\cite{FuWGLLJ24_app_health_network_19}, which demonstrates FGL's applicability in large-scale epidemic modeling.
\subsubsection{\textbf{Healthcare IoT and Monitoring}}
FGL is applied to edge-based elderly care~\cite{ghosh2023_app_health_network_11} and activity recognition in GraFeHTy~\cite{sarkar2021_app_health_network_8}. \cite{lou2021_app_health_network_3} and \cite{molaei2024_app_health_network_10} focus on processing temporal data from wearable monitoring devices.

The wide range of data modalities in FGL health applications highlights its versatility to further expand its influence from epidemic-surveillance knowledge graphs~\cite{wu2023_app_health_network_12}, disease transmission graph~\cite{Wan_EpiODE_ICML25}, graph-based epidemic modeling~\cite{Epi_Survey_KDD24} to malware detection in medical IoT devices~\cite{amjath2025_app_health_network_7}. 

\noindent 
\subsection{Computer Vision}
In computer vision, data is commonly represented as images or videos, which can be modeled as graphs where nodes denote visual features (e.g., objects, regions, pixels) and edges capture spatial or semantic relationships. FGL has supported progress in several directions by enabling decentralized training without compromising data privacy.
\subsubsection{\textbf{Architectural Innovations}}
\cite{lalitha2019_app_computer_vision_4} introduced a peer-to-peer learning framework for graph-structured vision data, extended by \cite{caldarola2021_app_computer_vision_2} with a cluster-driven approach for multi-domain recognition. \cite{he2023_app_computer_vision_1} proposed hierarchical model embedding for managing visual hierarchies, while \cite{litany2022_app_computer_vision_3} employed graph hypernetworks to align heterogeneous models.
\subsubsection{\textbf{Learning Paradigms}}
Theoretical foundations for multitask learning via Laplacian regularization were established by \cite{dinh2022_app_computer_vision_5}, and \cite{FedLIT} addressed latent link-type heterogeneity in visual relationship graphs, aiding tasks like scene understanding.

FGL enables collaborative training on visual tasks such as object detection and face recognition using distributed devices like smartphones and IoT cameras, without requiring centralized data. Key challenges remain in efficiently transmitting high-dimensional visual features and maintaining robustness under domain shifts, where graph-based methods offer natural advantages due to their ability to model inter-entity relationships.

\section{FGL in the era of PLMs}
\label{sec: PLM}
Pre-trained Large Models (PLMs) encompass a broad spectrum of generative, large-scale models trained on extensive datasets, with LLMs being the most prominent due to their widespread applications in diverse aspects of daily life. LLMs mark a pivotal advancement in ML, transitioning the field from purely analytical tools toward multifunctional agents capable of complex reasoning and, in some domains, substituting human experts. This success has inspired researchers to explore PLMs trained on various data modalities and to develop specialized models tailored to specific tasks. While we adopt PLMs as a general term for all major generative AI developments, the dominance of LLMs in recent literatures often necessitates using LLM-specific terminology when citing relevant works.

The collaborative nature of FL aligns well with the distributed demands of training and deploying such large models.~\cite{FederatedScope_LLM} By enabling secure and privacy-conscious utilization of globally distributed datasets, FL facilitates pre-training and domain-specific adaptation of PLMs without centralized data aggregation.~\cite{LLM_Future_Federated} Although the integration of FL with PLMs is still a nascent topic, it represents a critical frontier for advancing FGL and expanding its relevance in the era of Big Data, warranting further research to fully harness its potential.

\subsection{Existing PLM-FL Studies}

Several recent studies explore the integration of FL with Parameter-Efficient Fine-Tuning (PEFT) to adapt pre-trained LLMs to specialized tasks under resource and data constraints.
SLoRA~\cite{SLoRA} examines the impact of client heterogeneity on PEFT efficiency in federated settings. To mitigate this, it introduces a cost-effective fine-tuning strategy that achieves performance comparable to full fine-tuning.

FedBiOT~\cite{FedBiOT} tackles the limited accessibility of closed-source LLMs by training compressed LLM components locally using self-collected data from diverse domains. It employs PEFT to minimize communication overhead while leveraging knowledge distillation to approximate the performance of full-scale models.
FATE-LLM~\cite{FATE_LLM} presents a unified framework for fine-tuning LLMs in FL, enabling collaboration among isolated data holders using privacy-aware mechanisms.
FedPepTAO~\cite{FedPepTAO} focuses on prompt tuning within FL. It introduces a scoring mechanism to identify and selectively update critical prompt layers aligned with LLM outputs, optimizing both training efficiency and resource usage.
In a more foundational contribution, \cite{LLM_Future_Federated} demonstrates the feasibility of pre-training billion-parameter LLMs under heterogeneous federated conditions using the open-source \textit{Flower} framework~\cite{flower}.
\vspace{-0.3cm}
\subsection{Extending PLM-FL to PLM-FGL}
\label{subsec: PLM-FGL}
In recent years, researchers have initiated on integrating GML with PLMs substantially from the data-centric perspectives.
Graphs offer a natural way to represent complex relationships, and GNNs are well-suited for learning from such structures. Integrating these capabilities may expand the applicability of PLMs to non-textual domains, enabling more intelligent and structurally aware systems.
We highlight two strategies here.
The first tactic is to extract structural semantics from graph by leveraging GNNs, and then feed the graph-structured token sequences into PLMs to achieve alignments between graph topologies and natural language representations.~\cite{GraphGPT,HiGPT,UniGraph,GraphLLM,InstructMol}.
The second approach is to leverage PLMs' prior knowledge to supplement graph data, such as acquiring high-quality node embeddings or labels, to boost the downstream tasks.~\cite{G_prompt,OFA,TAPE,OpenGraph,RLMRec}. 

While research explicitly focused on PLM-FGL integration remains limited, its potential is considerable.
PLMs inherit surreal reasoning capabilities through the data-intense pre-training and has demonstrated by activating through few-shot~\cite{fewshot-GNNs} and zero-shot~\cite{zeroshot-GNNs} prompts.
Utilizing its reasoning abilities, researchers can explore PLMs's potential in mitigating the data heterogeneity, such as generating proxy datasets, which is critical to FGL's effectiveness.
Moreover, the integration with LLMs explores to break the barriers of multi-tasks and multi-domain learning, which can expand FGL's applicability and the scope of its training datasets.  
Challenges are how effective can the alignment between GML and PLMs proceed under the collaborative training with regulations against data transfer.   


Current representative works lay out two distinctive approach.
First is the PLM-enhanced FGL, represented by LLM4FGL~\cite{LLM4FGL}. It generate NLP-based prompts to inquiry the LLM, and use latter's output to assist downstream tasks.   
The second approach offers an alternative paradigm that combines the strength of cross-domain generalizability offered by PLMs and collaborative training architecture of FGL to break constraints of single-machine training of the former and mitigate the impracticablity of FGL when facing data and tasks heterogeneity.
Representative works include FedGFM+~\cite{FedGFM+}, which leverage LLMs to enhance local model training with complementary text-aware inputs, and leverage FGL training paradigm to collaboratively train the global GFM model.

\section{Future Works}
\label{sec: future works}
\subsection{New Learning Paradigm in FGL}
With the advent of big data, the AI community is increasingly focused on integrating continuously arriving data into trained models and enabling data removal to ensure privacy and regulatory compliance.
This section highlights two emerging paradigms in GML that are closely related to FGL but remain underexplored to enhance FGL’s practical applicability.
\subsubsection{\textbf{Continual Graph Learning (CGL) with FGL}} Incorporating data from new entities is essential but challenging due to the \textit{catastrophic forgetting} problem \cite{CCGL_1, CCGL_2, CCGL_3}, where GNNs lose performance on prior tasks when trained on new data. Most CGL methods assume centralized training \cite{CGL_1, CGL_2}. POWER~\cite{power} pioneers federated CGL by mitigating knowledge loss from conflicting client updates.

\subsubsection{\textbf{Graph Unlearning with FGL}} FGL must address data removal to comply with privacy regulations~\cite{GDPR, CCPA}. Centralized graph unlearning balances removing sensitive data with maintaining task performance~\cite{MEGU, GNNDelete, CGU}. Federated Graph Unlearning faces additional challenges: \textit{removal requests may affect overlapping subgraphs across clients, disrupting aggregation and degrading performance}. Entire client removals can further impair training and model generalizability. \cite{Subgraph_federated_unlearning} uses sampling and knowledge distillation for node removal but focuses on node classification. FedLU~\cite{FedLU} addresses knowledge graphs but lacks generalization to other graph types.

\subsubsection{\textbf{Open-wrold Graph Learning (OWGL)}} Assigning unlabeled nodes to unseen classes across entire datasets requires models to enhance their generalizability and improve boundary differentiation.~\cite{OWGL}
In realistic scenarios, class-wise information is often limited for one data center, and FGL appears to be an ideal choice for this task.
FGL is valued for its collaborative mechanism, which enables training a collective model from cross-domain datasets distributed across data centers without relying on meticulously designed modules.  Whether FGL can be introduced as a auxiliary training architecture to effectively support the objectives of OWGL while maintaining strong privacy preservation presents a inspiring research direction.

These nascent directions pose key challenges for FGL: improving client compatibility to reduce knowledge conflicts in continual learning and addressing data-centric issues in unlearning to meet privacy and regulatory demands.

\subsection{Multimodal Graph Learning (MGL) and FGL applicability}

MGL integrates heterogeneous data types, such as images, language sequences, and sensor readings, to construct complex representations for each entity, aiming to develop learning systems that are both generalizable and robust against increasing data heterogeneity.~\cite{MGL}
Within the context of FGL, integrating MGL is increasingly demanded by real-world scenarios. Here, we highlight three application domains for future endeavors.
\subsubsection{\textbf{Internet of Things (IoT)}}
Considering each device as an individual entity within a confined environment such as a household, it is natural that different devices capture diverse multimodal data. For example, surveillance cameras record video while sensor-based thermometers log temperature fluctuations numerically. In this context, FGL research seeks to optimize coordination among devices based on their topological relationships and to train models that capture personalized patterns from such heterogeneous data distributions.
\subsubsection{\textbf{Medical Information Processing}}
Diagnosing patients often requires integrating multimodal test outcomes such as CT scans represented as pixel data, doctors’ observations recorded in text, and patient bios stored as language sequences in medical systems. Unlike the IoT example where each node’s features originate from the same data source but exhibit multimodal heterogeneity across entities, in this medical context each patient’s features are derived from multiple data modalities. Furthermore, when medical services are distributed across distinct locations, this creates a vertical FGL scenario in which each client holds homogeneous data sources but significant heterogeneity exists between clients. 
Carefully crafted tasks and specific simulation strategies become paramount.
\subsubsection{\textbf{Recommendation System}}

Conventional FGL research in recommendation systems typically formulates node features based on textual data, such as user reviews and desensitized user biographies. Integrating multimodal data sources can further enrich user representations by incorporating posted images and behavioral logs that capture user preference, such as numbers of clicks. FGL stands to benefit significantly from these enhanced representations, which can lead to more accurate modeling and improved recommendation performance.

\subsection{New collaboration paradigm in the era of large model }

The integration of PLMs has sparked a revolution in training paradigms for GML, leveraging PLMs’ superior textual reasoning and analytical capabilities. Several surveys have explored promising synergies between GML and PLMs. 
FGL, as a distributed training framework addressing data silo challenges in conventional GML, is increasingly recognized as a potential collaborator with PLMs, and vice versa. We have discussed two existing literatures that have prioneered for such endeavors in Sec. ~\ref{subsec: PLM-FGL}, and this section intends to offer two generalizable collaboration paradigms that may inspire research aimed at enhancing the efficiency of FGL, PLMs, or both. 
The motivations are clear. We investigate roles of FGL in facilitating the implementation of PLMs into more challenging settings that meet the realistic demands, and role of PLMs in facilitate more robust and efficient FGL training paradigm.

\subsubsection{\textbf{FGL-enhanced PLMs paradigm}}

By combining FGL’s distributed training architecture and its proven ability to extract high-quality graph representations without sharing raw data, FGL can support PLMs training under stringent regulatory constraints. This enables PLMs to access diverse data samples across independent data centers. Notably, leveraging collective intelligence from independently trained small-scale PLMs facilitates a novel FGL training framework, allowing local training followed by model aggregation into larger PLMs, all while preserving privacy.
Moreover, graph can be utilized to introduce structural insights on uncoordinated raw data, particularly for multimodal source data. Partitioning data into domain-oriented clients and utilizing FGL for developing agent-based system can be a promising direction. 

\subsubsection{\textbf{PLMs-enhanced FGL paradigm}}
This paradigm addresses data-centric challenges in FGL, particularly those arising from data heterogeneity caused by incomplete entity attributes fragmented across data centers. The complementary knowledge and in-context reasoning capabilities of PLMs can significantly mitigate these limitations, serving as auxiliary inputs to enhance FGL training. 

\subsection{Communication and Privacy Preservation}

In the context of FGL, enhancing communication efficiency and maximizing privacy preservation during transmission are inherently synergistic goals. For instance, the prototype learning approach introduced in works such as FedPG~\cite{FedPG} simultaneously reduces communication overhead and improves security by utilizing proxy knowledge carriers.

Current research can be broadly classified into three focuses: optimizing learning efficiency while minimizing communication costs; enhancing communication stability within distributed, heterogeneous environments~\cite{lightweight, IoV, gossiping}; and prioritizing privacy preservation by refining differential privacy (DP) techniques widely adopted in FGL.

Extending these strategies to more challenging scenarios can drive the development of more robust and practical FGL methods. Key open questions include improving asynchronous communication among heterogeneous models, designing resilient communication architectures to mitigate network latency fluctuations, and exploring theoretical trade-offs between privacy protection and model utility to guide balanced implementations in real-world settings.

\subsection{Explainability}

Explainable Artificial Intelligence (XAI) exemplifies the growing societal demand for AI implementations that align with ethical and social principles. Within the context of FGL, XAI research primarily focuses on the design of aggregation strategies, with two notable approaches highlighted here.

\subsubsection{\textbf{Homophily Score-based}}
Methods, such as FedGTA~\cite{li2024fedgta} and FedTAD~\cite{zhu2024fedtad}, emphasize training efficiency by leveraging indices that measure the homophily of local subgraphs. These indices guide the iterative, interpretable selections of clients for participation in aggregation, thereby enhancing the trustworthiness of the process.

\subsubsection{\textbf{Shapley Value-based}}
The second approach utilizes the Shapley value, a game-theoretic concept widely applied in centralized XAI. Studies such as~\cite{IDS} employ Shapley values to identify significant attack patterns, bolstering defense robustness and privacy preservation;~\cite{recommendationX} use it to quantify feature contributions in recommendation system; and ~\cite{FLsamplingX} applies it for importance-based client selection. However, these methods lack further exploration within the FGL domain.

The aforementioned techniques contribute significantly to enhancing interpretability in FGL outcomes and mitigating issues such as data heterogeneity. Nonetheless, considerable opportunities remain to advance explainable FGL. Topics such as fairness learning and promoting equity across demographic groups, particularly through the integration of FGL with social science experiments, warrant sustained research efforts.

\section*{Conclusion}

This survey reorganizes existing FGL studies through a data-centric lens using a two-level taxonomy that aligns with recent trends in data-centric graph machine learning. It systematically highlights the structural and distributional properties of datasets (Data Characteristics) and the training strategies used to address data-related challenges (Data Utilization), each comprised of three orthogonal criteria.
The taxonomy serves as a practical reference, enabling researchers to quickly identify works relevant to their desired data-centric objectives. 
In addition, this survey extends its scope to explore the integration of FGL with large language models, which increasingly contribute to the enrichment and interpretability of graph-structured data, further reinforcing the data-centric viewpoint. 
We further explore opportunities to extend FGL into emerging GML domains, promote multimodal data collaboration, and propose two potential training paradigms that leverage the strengths of both FGL and PLMs. 
Additionally, we review advancements in optimizing communication efficiency and privacy preservation, offer recommendations for improving interpretability in FGL, and discuss pathways to enhance its practical deployment across real-world applications.
By advancing FGL’s applicability and encouraging the extension to diverse research domains and increasingly complex datasets, we envision a future where FGL not only overcomes current limitations but also catalyzes transformative developments across multidisciplinary fields.


\bibliographystyle{IEEEtran}
\bibliography{ref}

\end{document}